\documentclass[sigconf,nonacm]{acmart}

\AtBeginDocument{%
  }

\usepackage{multirow}
\usepackage{graphicx}
\usepackage{subcaption}
\usepackage{stfloats}

\begin{document}

\title{FractalForensics: Proactive Deepfake Detection and Localization via Fractal Watermarks}

\author{Tianyi Wang}
\orcid{0000-0003-2920-6099}
\email{wangty@nus.edu.sg}
\affiliation{
  \institution{National University of Singapore}
  \city{Singapore}
  \country{Singapore}
}

\author{Harry Cheng}
\orcid{0000-0001-7436-0162}
\email{xaCheng1996@gmail.com}
\affiliation{%
  \institution{National University of Singapore}
  \city{Singapore}
  \country{Singapore}
}

\author{Ming-Hui Liu}
\email{liuminghui@mail.sdu.edu.cn}
\affiliation{%
 \institution{Shandong University}
 \city{Jinan}
 \state{Shandong}
 \country{China}
}

\author{Mohan Kankanhalli}
\email{mohan@comp.nus.edu.sg}
\affiliation{
  \institution{National University of Singapore}
  \city{Singapore}
  \country{Singapore}
}
\authornote{Corresponding author.}

\begin{abstract}
Proactive Deepfake detection via robust watermarks has seen interest ever since passive Deepfake detectors encountered challenges in identifying high-quality synthetic images. However, while demonstrating reasonable detection performance, they lack localization functionality and explainability in detection results. Additionally, the unstable robustness of watermarks can significantly affect the detection performance. In this study, we propose novel fractal watermarks for proactive Deepfake detection and localization, namely \textit{FractalForensics}. Benefiting from the characteristics of fractals, we devise a parameter-driven watermark generation pipeline that derives fractal-based watermarks and performs one-way encryption of the selected parameters. Subsequently, we propose a semi-fragile watermarking framework for watermark embedding and recovery, trained to be robust against benign image processing operations and fragile when facing Deepfake manipulations in a black-box setting. Moreover, we introduce an entry-to-patch strategy that implicitly embeds the watermark matrix entries into image patches at corresponding positions, achieving localization of Deepfake manipulations. Extensive experiments demonstrate satisfactory robustness and fragility of our approach against common image processing operations and Deepfake manipulations, outperforming state-of-the-art semi-fragile watermarking algorithms and passive detectors for Deepfake detection. Furthermore, by highlighting the areas manipulated, our method provides explainability for the proactive Deepfake detection results\footnote{Code will be available at \url{https://github.com/wangty1/fractalforensics}.}. 
\end{abstract}



\keywords{Deepfake Detection, Deepfake Manipulation Localization, Semi-Fragile Watermarks, Digital Forensics}


\maketitle

\section{Introduction}

The rapid development of Deepfake image manipulation techniques has introduced significant risks to individuals’ cyber privacy and safety, despite offering benefits to society in areas such as entertainment and education~\cite{DeepfakeSurvey[20]}. Since passive Deepfake detection methods~\cite{FFPP[16],luo2023beyond[56],SBIs[17],kong2024moe[54],CADDM[19],RECCE[18],luo2024forgery[55]} often face performance bottlenecks when encountering previously unseen generative algorithms, recent studies have begun to explore proactive defenses that intervene prior to image tampering.

\begin{figure*}
\centering
\includegraphics[width=0.99\textwidth]{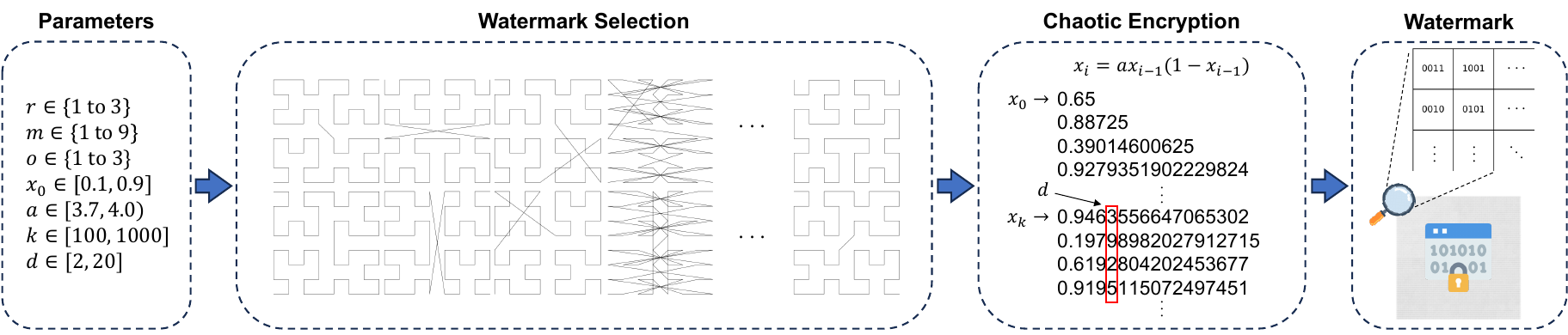}
\caption{Pipeline of watermark generation and encryption based on user-chosen parameters. Given a standard Hilbert curve, parameters $r$, $m$, and $o$ determine rotation, mirroring, and order modification variations when deriving the raw watermark, while $x_0$, $a$, $k$, and $d$ determine how the watermark is securely encrypted in the chaotic system. }
\label{fig:encryption_pipeline}
\end{figure*}

Proactive defense using watermarks can be broadly categorized into robust and semi-fragile methods. Although both aim to insert predefined watermarks for image protection and ensure robustness under common image processing operations, the former is designed to maintain robustness even under malicious Deepfake manipulations~\cite{IDPMark2024Wang[13],LampMark2024Wang[12]}, while the latter intentionally allows the watermarks to degrade~\cite{FaceGuard[23],FaceSigns[3]}. However, existing methods face several key limitations. First, to consistently retain watermark integrity, robust watermarking frameworks usually require training with Deepfake manipulations. Therefore, the robustness tends to significantly degrade when encountering unseen Deepfake manipulation categories\footnote{E.g., LampMark~\cite{LampMark2024Wang[12]} has an average of 96.46\% watermark recovery rate against face swapping algorithms (trained with) but an average of 73.61\% against face reenactment algorithms (unseen). }, undermining the reliability of detection results. On the other hand, while allowing fragility when recovering watermarks from Deepfake-manipulated images, the semi-fragile approaches are usually vulnerable to common image processing operations that imitate real-life scenarios. Furthermore, to support forensic investigations in practical Deepfake-related criminal cases, beyond Deepfake detection, the current methods lack explainable localizations of the Deepfake-manipulated area within suspect images. Lastly, since most semi-fragile methods mainly focus on the watermarking frameworks, the watermark confidentiality and ease of use have been barely discussed. 

To address the aforementioned issues, we propose FractalForensics, a semi-fragile watermarking approach for proactive Deepfake detection and localization. First, we design a parameter-driven watermark generation pipeline (Figure~\ref{fig:encryption_pipeline}). Specifically, to achieve the localization functionality, we use parameterized fractal curves~\cite{Allezaud_1985[24]} to generate watermarks with spatially meaningful structures and minimal storage overhead. Unlike randomly defined watermark patterns, we devise fractal-based watermarks that are generated following deterministic rules and a small set of parameters, allowing on-demand reconstruction without storing ground-truth layouts. Meanwhile, the recursive and space-filling nature of fractals ensures that watermarks span across most regions of images, which is essential for localizing Deepfake-manipulated areas. On top of the structural backbone, we apply variations to diversify fractal watermark shapes, and to guarantee watermark confidentiality, we further utilize a chaotic encryption system~\cite{ChaoticEnc[25]} to perform one-way encryption on fractal watermarks, ensuring unpredictability and irreversibility. In general, the entire pipeline is parameter-driven and confidential by design, eliminating the need to store raw watermark data. 

Then, we construct an end-to-end watermarking framework for embedding and recovery (Figure~\ref{fig:workflow}). To preserve spatial correspondence and enable localization, each fractal watermark is represented as a 2-dimensional matrix with 4-bit entries, each encoded as a 4-channel binary tensor. We introduce an entry-to-patch watermark embedding strategy that maps each watermark entry to the image patch at the corresponding position. In particular, for the input image, an image feature extraction module obtains critical image information and a watermark diffusion module expands and refines the input watermark, where outputs of the two modules are passed to the watermark fusion module for watermarked image reconstruction. Thereafter, we construct a decoder for watermark recovery with expected robustness and fragility. In the end, proactive Deepfake detection and localization are accomplished based on the recovered watermarks. The entire framework is trained in a black-box condition without Deepfake manipulations. Extensive experiments further establish the satisfactory Deepfake detection and localization ability of the proposed FractalForensics. The contributions of this work are threefold:
\begin{itemize}
\item We propose FractalForensics, a novel watermarking framework via semi-fragile fractal watermarks for proactive Deepfake detection and localization in the black-box scenario. 
\item Based on the characteristics of fractals, we devise a parameter-driven watermark generation pipeline and introduce an entry-to-patch watermarking strategy that implicitly ensures explainable localization of the manipulated area.
\item Experiments demonstrate promising robustness and fragility of our watermarks under common and Deepfake manipulations, outperforming state-of-the-art passive and semi-fragile approaches. Additionally, our method generalizes well to unseen datasets and Deepfake manipulations.
\end{itemize}

\section{Related Work}

\subsection{Watermarks for Deepfake Detection}

Although classic passive detectors~\cite{FFPP[16],NoiseDF[22],SBIs[17],DCPT[21],kong2022detect[52],kong2025pixel[53]} have achieved significant progress since the first occurrence of Deepfake, recent advancements in generative models~\cite{APSwap[26],UniFace[6],DiffSwap[8],tan2024rle[51]} have brought strong challenges with high-quality synthetic images. As countermoves, proactive approaches~\cite{CMUA[28],DFRAP[29],NullSwap2025Wang[14],SepMark[2],IDPMark2024Wang[13]} insert invisible signals into benign images in advance of Deepfake manipulations and actively determine real or fake based on the signals' functionality afterwards. 

\begin{figure*}
    \centering
    \includegraphics[width=\textwidth]{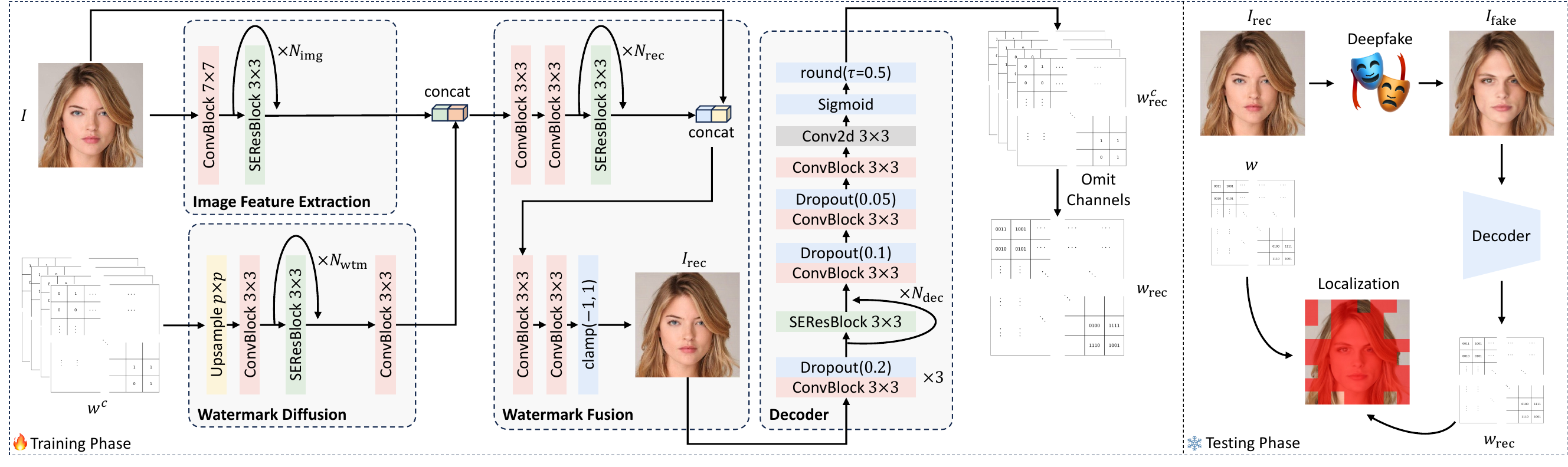}
    \caption{Workflow of the proposed FractalForensics. The channel-wise watermark matrix $w^c$ is embedded into image $I$ following image feature extraction, watermark diffusion, and watermark fusion modules. To recover the watermark from $I_\textrm{rec}$, we construct a decoder. The framework is trained end-to-end. In the testing phase, Deepfake localization is implicitly achieved. }
    \label{fig:workflow}
\end{figure*}

Robust watermarking frameworks~\cite{HiDDeN[33],MBRS[32],CIN[34]} are traditionally used for copyright verification and authentication, while recent work assigns semantics to the robust watermarks for Deepfake detection. In 2021, FakeTagger~\cite{Wang2021FakeTagger[31]} raises the general idea of embedding robust and meaningful messages into images for proactive Deepfake detection and provenance. Wang \textit{et al.}~\cite{IDPMark2024Wang[13]} proposed identity perceptual watermarks that are robustly embedded and recovered for the detection of Deepfake face swapping and source tracing based on consistency in identity information. Later, LampMark~\cite{LampMark2024Wang[12]} exploits structure-sensitive characteristics and introduces landmark perceptual watermarks, performing proactive Deepfake detection for both face swapping and face reenactment. However, being trained only against the face swapping model, SimSwap~\cite{SimSwap[15]}, its robustness drops significantly when evaluated against unseen face reenactment algorithms. 

Semi-fragile watermarks are designed to be robust and fragile regarding common image processing operations and Deepfake manipulations, respectively. An early work, FaceGuard~\cite{FaceGuard[23]}, raises the idea of assigning fragility to watermarks when facing Deepfake manipulations. Later, Kong et al.~\cite{kong2022detect[52]} proposed an innovative dual-branch architecture and achieved state-of-the-art performance in Deepfake localization. WaterLo~\cite{WaterLo[1]} uniformly applies a 1-bit message to the entire benign image and achieves localization for early face swapping algorithms depending on the presence of watermarks in the facial area. Zhao \textit{et al.}~\cite{Zhao2023WACV[36]} inserted a watermark into the identity features of the images, and upon modifying facial identities by face swapping manipulations, the image authenticity is determined based on the absence of the watermark. Similarly, FaceSigns~\cite{FaceSigns[3]} embeds predefined 128-bit semi-fragile binary watermarks into images via a UNet architecture for Deepfake detection. Recently, EditGuard~\cite{EditGuard[4]} and OmniGuard~\cite{OmniGuard[37]} treat this task as a joint image-bit steganography issue and embed semi-fragile watermarks that accomplish detection and localization regarding AI-generated contents (AIGC) simultaneously. Additionally, some recent studies~\cite{sun2024faketracer[38],sun2022faketracer[39],SepMark[2]} propose to utilize robust and semi-fragile watermarks together and achieve proactive detection based on the co-existence of the watermarks, where their semi-fragile watermarks are also worth comparison and discussion in this study. 

\subsection{Fractals Patterns for Structural Coverage}

Fractals~\cite{Allezaud_1985[24]} have been widely studied for their space-filling properties to traverse all points in a 2-dimensional grid~\cite{SaganFractals1994[40]}. While exhibiting strong spatial continuity, self-similarity, and locality preservation, these curves have been proven to be suitable for applications in image compression~\cite{HilbertImageCompression1999[41]} and texture synthesis~\cite{PeanoTexture2011[42]}. However, their potential for structured watermark design in the context of proactive Deepfake detection remains largely unexplored. In this work, leveraging the fractal patterns, we derive parameter-driven watermark matrices that are spatially coherent, enabling implicit localization besides proactive Deepfake detection. 

\section{Methodology}

\subsection{Overview}

Deepfakes manipulate facial areas in full image frames. To protect the faces, we propose FractalForensics, a semi-fragile watermarking framework for proactive Deepfake detection and localization. For a clean facial image $I$ cropped from a full image frame, an encrypted fractal watermark matrix $w$ is generated via a parameter-driven pipeline (Figure~\ref{fig:encryption_pipeline}). This pipeline incorporates user-selected parameters that define the fractal type, structural variation, and chaotic encryption keys. Subsequently, a watermarking framework (Figure~\ref{fig:workflow}) is established to invisibly embed $w$ into $I$, consisting of an image feature extraction module, a watermark diffusion module, and a watermark fusion module, to produce the watermarked image $I_\textrm{rec}$. A decoder is then constructed to recover the watermark from $I_\textrm{rec}$ as $w_\textrm{rec}$. By comparing $w_\textrm{rec}$ with the original $w$, we achieve proactive Deepfake detection and localization. 

\subsection{Watermark Generation Pipeline}

Fractals are recursive geometric shapes whose structures are derived based on parameters. Rather than storing all watermarks like in previous studies, we propose a watermark generation pipeline that produces encrypted watermarks depending on the selected parameters. In this study, we use the Hilbert curve~\cite{Hilbert1891[43]} as the selected fractal shape for demonstration. The Hilbert curve of order $n$ defines a continuous path that maps each integer index $i \in [0, 2^{2n} - 1]$ to a unique coordinate in a $2^n \times 2^n$ grid. The mapping follows a recursive and self-similar rule that fills the space while preserving spatial locality. Relying on the traversal order defined by the standard Hilbert curve, we construct a 2-dimensional watermark matrix as a representation of the curve. 

To ensure diversity in the selected fractal shapes, we apply shape variations toward the standard Hilbert curve by rotations $r \in$ \{\textrm{0: 0°}, \textrm{1: 90°}, \textrm{2: 180°}, \textrm{3: 270°}\}, mirroring \texttt{$m \in$ \{\textrm{0: None}, \textrm{1: Top}, \textrm{2: Bottom}, \textrm{3: Left}, \textrm{4: Right}, \textrm{5: Top Left}, \textrm{6: Top Right}, \textrm{7: Bottom Left}, \textrm{8: Bottom Right}\}}, and order modifications \texttt{$o \in$ \{\textrm{0: None}, \textrm{1: Reverse}, \textrm{2: Zigzag}, \textrm{3: Cross Flip}\}}, leading to 144 different unique shape variations based on the combination of parameters. For each curve, there is a start and an end point that record the order and direction. To securely protect the curve shape along with the order and direction, we utilize a chaotic encryption~\cite{ChaoticEnc[25]} system to perform one-way encryption based on the selected parameters. In specific, the logistic map is defined by 
\begin{equation}
x_i = ax_{i-1}(1-x_{i-1}),
\label{eq:logistic_map}
\end{equation}
where $x_0 \in [0.1, 0.9]^\mathbb{R}$ and $a \in [3.7, 4.0)^\mathbb{R}$ are parameters that determine the chaotic map. To ensure the unpredictability and irreversibility of encrypted watermarks, we let parameters $k \in [100, 1000]^\mathbb{Z}$ and $d \in [2, 20]^\mathbb{Z}$ determine the starting value $x_k$ of the chaotic sequence and the index of the selected digit of each value. We use the $2^{2n}$ consecutive values at the $d$-th digit starting from $x_k$ as the encrypted watermark matrix of the raw entries following the curve shape from the start to the end point. Due to the chaotic characteristic of the encryption system, unless the confidential parameters are known, attempting to recover the original watermark curve by brute force is worse than NP-Hard~\cite{NP-Hard[50]}. For watermarking purposes, since there is more tolerance for binary entries when encoding and decoding, the encrypted watermark entries are transformed into binary formats with 4 bits for each entry. 

\subsection{Watermarking Framework}

We construct a watermarking framework with deep neural networks, aiming to invisibly embed watermark $w$ into image $I$ and perform recovery with robustness and fragility accordingly. To achieve the localization functionality for tampered areas, we introduce an innovative entry-to-patch strategy for watermarking. Specifically, the input image $I$ is implicitly recognized as patches of size $32\times32$, and the $2^n \times 2^n$ patches are positionally matched to the watermark matrix $w$ of the same size. To ensure locality preservation such that the watermarks are positionally embedded in the desired locations, we place the 4-bit entries on the channel dimension so that they share the same spatial positions, resulting in the channel-wise binary watermark matrix $w^c$ of shape $4 \times 2^n \times 2^n$.

\noindent\textbf{Image Feature Extraction.} We construct an image feature extraction module that contains a ConvBlock with a kernel size of $7 \times 7$ followed by $N_\textrm{img}$ SEResBlocks with kernel sizes of $3 \times 3$. In particular, a ConvBlock contains a sequentially executed CNN layer, batch normalization layer, and LeakyReLU activation. An SEResBlock utilizes the ResNet~\cite{ResNet[44]} bottleneck block and applies the squeeze-and-excitation networks~\cite{SENet[45]} (SENet) for efficient channel-wise attention. For image $I$, we derive critical image features, with residually preserved information from early stages by SEResBlocks, for image reconstruction in the later module. The image features are analyzed in the pixel-aligned feature domain such that spatial dimensions of intermediate image features remain unchanged to maintain strict position alignment, providing favorable conditions for entry-to-patch watermarking with minimal information losses.

\noindent\textbf{Watermark Diffusion.} To enable effective and robust watermark embedding, we expand and refine the binary channel-wise watermark matrix $w^c \in \{0,1\}^{4 \times 2^n \times 2^n}$ before fusion for image reconstruction. In specific, the watermark diffusion module first upsamples $w^c$ to the same spatial dimension as the input image, where each 4-bit entry is spatially expanded into the corresponding $32 \times 32$ patch. Thereafter, the upsampled watermark is fed to a ConvBlock, $N_\textrm{wtm}$ SEResBlocks, and a ConvBlock to adaptively refine locally critical watermark features via moderate kernel sizes, ensuring the robustness and invisibility of watermark signals during embedding. 

\noindent\textbf{Watermark Fusion.} In the watermark fusion module, we integrate the image and watermark features to produce the watermarked image $I_\textrm{rec}$. The two features are first concatenated along the channel dimension and sequentially passed through two ConvBlocks and $N_\textrm{rec}$ SEResBlocks, to accomplish a feature-level reconstruction while maintaining the entry-to-patch correlation locally via moderate kernel sizes and merely channel-wise attention operations, avoiding over-coherence between patches that are far apart since this can make the watermark overly robust against Deepfake when watermark fragility is required. Dropout operations are concurrently applied to avoid possible overfitting and unexpectedly robust watermark recovery. Then, to enhance reconstruction quality and obtain the ultimate RGB image, we concatenate the reconstructed feature-level result with the original image $I$ and pass them through two consecutive ConvBlocks to generate the watermarked image $I_\textrm{rec}$ with three channels. 

\noindent\textbf{Decoder.} A lightweight decoder is designed for watermark recovery. In specific, a sequence of a ConvBlock, $N_\textrm{dec}$ SEResBlocks, and two ConvBlocks that gradually reduce spatial resolution and distill deterministic watermark information are applied to the watermarked image $I_\textrm{rec}$. With moderate kernel sizes, the decoder preserves positional correspondence regarding the entry-to-patch watermarking strategy, and stride-based downsampling is employed instead of pooling operations. A final Sigmoid activation followed by rounding is applied to map the output to binary values. In this end-to-end pipeline, the decoder is jointly trained with the former modules for watermark embedding, which benefits the thorough learning of the entry-to-patch scheme and balances the performance regarding visual quality and watermark robustness. 

\noindent\textbf{Discriminator.} In this work, a discriminator is trained from scratch along with the main FractalForensics framework. Specifically, a ConvBlock and $N_\textrm{rec}$ SEResBlocks are applied simply followed by a binary classifier to determine whether it is watermarked or not. The discriminator is used to train the main framework adversarially. 

\begin{table}
\caption{Visual quality evaluation of the watermarked images. Information includes model name, resolution (Res.), watermark size (Num. Bits), PSNR, SSIM, and LPIPS. }
\label{tab:quantitative_common_wtm}
\begin{tabular}{lccccc}
\toprule
Model & Res. & Num. Bits & PSNR$\uparrow$ & SSIM$\uparrow$ & LPIPS$\downarrow$ \\
\midrule
WaterLo~\cite{WaterLo[1]} & 512 & 1 & \textbf{46.73} & \textbf{0.993} & \textbf{0.001} \\
\multirow{2}{*}{SepMark~\cite{SepMark[2]}} & 128 & 30 & 31.51 & 0.965 & 0.014 \\
 & 256 & 128 & 32.08 & 0.945 & 0.046\\
FaceSigns~\cite{FaceSigns[3]} & 256 & 128 & 30.60 & 0.945 & 0.033 \\
EditGuard~\cite{EditGuard[4]} & 512 & 64 & 36.93 & 0.841 & 0.021 \\
\multirow{2}{*}{FractalForensics} & 256 & 256 & 35.926 & 0.974 & \underline{0.013} \\
 & 512 & 1024 & \underline{38.129} & \underline{0.981} & 0.014 \\
\bottomrule
\end{tabular}
\end{table}

\subsection{Loss Functions}

\begin{table*}[t!]
\centering
\caption{Watermark recovery rate against common image operations on CelebA-HQ. }
\resizebox{\textwidth}{!}{
\begin{tabular}{lccccccc|cc}
\toprule
Models & WaterLo~\cite{WaterLo[1]} & \multicolumn{2}{c}{SepMark~\cite{SepMark[2]}} & FaceSigns~\cite{FaceSigns[3]} & EditGuard~\cite{EditGuard[4]} & \multicolumn{2}{c|}{FractalForensics (patch)} & \multicolumn{2}{c}{FractalForensics (bit)} \\
\midrule
Resolution & 512 & 128 & 256 & 256 & 256 & 256 & 512 & 256 & 512 \\
\midrule
Identity & 99.99\% & 99.99\% & 99.99\% & 99.84\% & 99.98\% & 99.99\% & 99.99\% & 99.99\% & 99.99\% \\
Jpeg & 61.72\% & 99.93\% & 99.93\% & 96.68\% & 80.01\% & 99.97\% & 99.84\% & 99.99\% & 99.95\% \\
Gaussian Noise & 64.20\% & 54.47\% & 49.38\% & 52.08\% & 58.39\% & 99.76\% & 99.04\% & 99.93\% & 99.68\% \\
Gaussian Blur & 61.76\% & 99.99\% & 99.99\% & 98.52\% & 0.23\% & 99.57\% & 99.96\% & 99.88\% & 99.99\% \\
Median Blur & 89.25\% & 99.99\% & 99.99\% & 98.44\% & 3.86\% & 99.99\% & 99.97\% & 99.96\% & 99.99\% \\
Resize & 61.70\% & 99.99\% & 99.99\% & 96.73\% & 3.54\% & 98.03\% & 99.56\% & 99.47\% & 99.86\% \\
\midrule
Average$\uparrow$ & 73.10\% & 92.39\% & 91.55\% & 90.38\% & 41.00\% & \underline{99.55\%} & \textbf{99.73\%} & 99.87\% & 99.91\% \\
\bottomrule
\end{tabular}
}
\label{tab:common_recovery_rate}
\end{table*}

In this study, we focus on two main objectives: visual quality and watermark recovery performance. We first assign a pixel-level Mean Squared Error (MSE) loss on $I$ and $I_\textrm{rec}$ by
\begin{equation}
\mathcal{L_\textrm{MSE}} = \lVert I - I_\textrm{rec} \rVert^2_2.
\end{equation}
Meanwhile, to make sure that $I_\textrm{rec}$ is perceptually identical to $I$ for human eyes, we adopt the Learned Perceptual Image Patch Similarity (LPIPS) loss following
\begin{equation}
\mathcal{L_\textrm{LPIPS}} = \Sigma_l \lVert \phi_l(I) - \phi_l(I_\textrm{rec}) \rVert_2^2,
\end{equation}
where $\phi_l$ represents the feature activation from layer $l$ of a pre-trained AlexNet~\cite{AlexNet[46]}. Meanwhile, the discriminator $D$ is trained to distinguish $I$ and $I_\textrm{rec}$ following
\begin{equation}
\mathcal{L}_D = -\mathbb{E}(\log D(I)) +\mathbb{E}\log(1-D(I_\textrm{rec})),
\end{equation}
and it assists in training the watermarking framework for better visual quality following
\begin{equation}
\mathcal{L}_\textrm{adv} = -\mathbb{E}(\log D(I_\textrm{rec})).
\end{equation}
We train the decoder for watermark robustness following 
\begin{equation}
\mathcal{L}_\textrm{dec} = \lVert m_\textrm{rec} - m \rVert.
\end{equation}
In summary, the total loss is computed following 
\begin{equation}
\mathcal{L} = \lambda_\textrm{MSE}\mathcal{L_\textrm{MSE}} + \lambda_\textrm{LPIPS}\mathcal{L_\textrm{LPIPS}} + \lambda_\textrm{adv}\mathcal{L}_\textrm{adv} + \lambda_\textrm{dec}\mathcal{L}_\textrm{dec}.
\end{equation}

\section{Experiments}

\subsection{Implementation Details}

We used the popular facial image datasets, CelebA-HQ~\cite{CelebAHQ[47]} and LFW~\cite{LFW[48]}, for experiments. Specifically, we followed the official split of CelebA-HQ with 30,000 facial images for training, validation, and testing, and we used LFW with 5,749 facial identities for evaluating the generalizability. We trained two versions of FractalForensics for the 256 and 512 resolutions, with $N_\textrm{img}$ being 5 and 6, $N_\textrm{wtm}$ being 5 and 4, $N_\textrm{rec}$ being 5 and 4, and $N_\textrm{dec}$ being 4 and 3, respectively. For the loss weights, we set $\lambda_\textrm{MSE}=1$, $\lambda_\textrm{LPIPS}=0.5$, $\lambda_\textrm{adv}=0.01$, and $\lambda_\textrm{dec}=12$. During training, we used Identity, Jpeg, and Gaussian Noise as common image processing operations to boost watermark robustness. Deepfake manipulations are completely omitted during training under the black-box setting. In the testing phase, the semi-fragile watermarking algorithms are evaluated against common image processing operations including Identity, Jpeg, Gaussian Noise, Gaussian Blur, Median Blur, and Resize and Deepfake manipulations including SimSwap~\cite{SimSwap[15]}, InfoSwap~\cite{InfoSwap[5]}, UniFace~\cite{UniFace[6]}, E4S~\cite{E4S[7]}, and DiffSwap~\cite{DiffSwap[8]} for face swapping and StarGAN~\cite{StarGAN-V2[9]}, StyleMask~\cite{StyleMask[10]}, and HyperReenact~\cite{HyperReenact[11]} for face reenactment. We trained the main watermarking framework at a learning rate of $0.002$ and trained the discriminator at $0.0004$. All experiments are conducted on 8 Tesla A100 GPUs. 

\subsection{Watermarking Performance on CelebA-HQ}

In this section, we evaluate the watermarking performance on CelebA-HQ~\cite{CelebAHQ[47]}, with respect to the visual quality of the watermarked images and the watermark robustness and fragility with respect to common image operations and Deepfake manipulations. 

\noindent\textbf{Visual Quality.} We conducted quantitative evaluations of the visual quality after embedding watermarks. Specifically, we compared our proposed FractalForensics to state-of-the-art semi-fragile watermarks for proactive defense against Deepfake and AIGC. As Table~\ref{tab:quantitative_common_wtm} reports, our method achieves the second-best performance with the most difficult bit-per-pixel rate, while the best WaterLo~\cite{WaterLo[1]} has watermarks with only 1 bit. In general, while all methods demonstrate reasonable LPIPS values, SepMark\footnote{SepMark consists of a robust and a semi-fragile watermark, where we adopted the semi-fragile watermark for comparing robustness and fragility in this study.}~\cite{SepMark[2]} and FaceSigns~\cite{FaceSigns[3]} both have PSNR values below 35, and EditGuard~\cite{EditGuard[4]} derives an unsatisfactory SSIM value below 0.85. As a result, our approach consistently maintains promising visual qualities at both resolutions for all evaluation metrics on visual quality. 

\begin{table*}[t!]
\centering
\caption{Watermark recovery rate against Deepfake image manipulations on CelebA-HQ. }
\begin{tabular}{lcccccccc}
\toprule
Models & WaterLo~\cite{WaterLo[1]} & \multicolumn{2}{c}{SepMark~\cite{SepMark[2]}} & FaceSigns~\cite{FaceSigns[3]} & EditGuard~\cite{EditGuard[4]} & \multicolumn{2}{c}{FractalForensics} \\
\midrule
Resolution & 512 & 128 & 256 & 256 & 256 & 256 & 512 \\
\midrule
SimSwap~\cite{SimSwap[15]} & 61.72\% & 98.97\% & 69.49\% & 50.21\% & 49.23\% & 78.19\% & 60.79\% \\
InfoSwap~\cite{InfoSwap[5]} & 62.10\% & 99.94\% & 98.38\% & 50.46\% & 37.04\% & 75.12\% & 72.55\% \\
UniFace~\cite{UniFace[6]} & 61.74\% & 50.11\% & 49.86\% & 49.95\% & 49.18\% & 64.66\% & 24.11\% \\
E4S~\cite{E4S[7]} & 62.34\% & 99.78\% & 95.33\% & 50.19\% & 50.05\% & 69.19\% & 66.81\% \\
DiffSwap~\cite{DiffSwap[8]} & 62.08\% & 50.23\% & 49.79\% & 51.79\% & 50.73\% & 81.31\% & 59.91\% \\
StarGAN~\cite{StarGAN-V2[9]} & 60.89\% & 49.93\% & 50.07\% & 50.17\% & 49.90\% & 9.92\% & 10.02\% \\ 
StyleMask~\cite{StyleMask[10]} & 66.27\% & 49.86\% & 50.10\% & 49.98\% & 50.14\% & 10.00\% & 9.96\% \\
HyperReenact~\cite{HyperReenact[11]} & 61.88\% & 50.11\% & 50.06\% & 50.13\% & 49.87\% & 9.98\% & 10.02\% \\
\midrule
Average$\downarrow$ & 62.47\% & 64.28\% & 63.37\% & 50.38\% & \underline{48.13\%} & 50.05\% & \textbf{39.27\%} \\
\bottomrule
\end{tabular}
\label{tab:fake_recovery_rate}
\end{table*}

\begin{table*}[t!]
\centering
\caption{Deepfake detection performance in AUC scores on CelebA-HQ. }
\resizebox{\textwidth}{!}{
\begin{tabular}{lcccc|ccccc}
\toprule
Model & Xception~\cite{FFPP[16]} & SBIs~\cite{SBIs[17]} & RECCE~\cite{RECCE[18]} & CADDM~\cite{CADDM[19]} & WaterLo~\cite{WaterLo[1]} & SepMark~\cite{SepMark[2]} & FaceSigns~\cite{FaceSigns[3]} & EditGuard~\cite{EditGuard[4]} & FractalForensics \\
\midrule
SimSwap~\cite{SimSwap[15]} & 71.15\% & 88.94\% & 69.01\% & 87.66\% & 69.77\% & 99.97\% & 98.87\% & 49.33\% & 99.99\% \\
InfoSwap~\cite{InfoSwap[5]} & 65.50\% & 80.50\% & 52.13\% & 61.39\% & 51.30\% & 64.83\% & 98.87\% & 50.36\% & 99.99\% \\
UniFace~\cite{UniFace[6]} & 70.34\% & 79.41\% & 67.35\% & 82.73\% & 69.69\% & 99.99\% & 98.91\% & 46.71\% & 99.99\% \\
E4S~\cite{E4S[7]} & 53.70\% & 61.05\% & 47.19\% & 73.13\% & 63.99\% & 73.12\% & 98.87\% & 56.37\% & 99.99\% \\
DiffSwap~\cite{DiffSwap[8]} & 53.62\% & 56.15\% & 63.60\% & 73.33\% & 52.03\% & 99.99\% & 98.83\% & 48.56\% & 99.99\% \\
StarGAN~\cite{StarGAN-V2[9]} & 49.30\% & 65.86\% & 41.55\% & 44.34\% & 68.56\% & 99.99\% & 98.87\% & 47.12\% & 99.99\% \\
StyleMask~\cite{StyleMask[10]} & 40.23\% & 48.45\% & 23.87\% & 39.73\% & 50.12\% & 99.99\% & 98.87\% & 47.03\% & 99.99\% \\
HyperReenact~\cite{HyperReenact[11]} & 76.27\% & 53.35\% & 78.23\% & 42.87\% & 64.86\% & 99.99\% & 98.88\% & 47.03\% & 99.99\% \\ 
\midrule
Average & 60.01\% & 66.71\% & 55.37\% & 63.15\% & 61.29\% & 92.23\% & \underline{98.87\%} & 49.06\% & \textbf{99.99\%} \\
\bottomrule
\end{tabular}
}
\label{tab:deepfake_detection}
\end{table*}

\noindent\textbf{Watermark Recovery Rate.} We evaluated the watermark recovery rate of our approach against common image processing operations and Deepfake manipulations, compared to contrastive semi-fragile watermarking frameworks. As demonstrated in Table~\ref{tab:common_recovery_rate}, while all algorithms stay robust as expected when applied with no operation (Identity), they each suffer a drastic reduction in watermark recovery rates when facing some image processing operations. In general, Gaussian Noise has the most serious effect, causing all watermark recovery rates to fall below 65\% for the comparative algorithms. Meanwhile, the two models that have demonstrated relatively better visual quality in Table~\ref{tab:quantitative_common_wtm}, WaterLo~\cite{WaterLo[1]} and EditGuard~\cite{EditGuard[4]}, also fail to maintain robustness regarding Jpeg, Gaussian Blur, Median Blur, and Resize operations. For our proposed FractalForensics, since a 4-bit binary value represents each watermark entry, we computed the patch-wise watermark recovery rate as our official performance and used that for Deepfake detection and localization in later sections, which is harder than the bit-wise metric in preserving robustness. As a result, our method achieves the best performance on average with 99.55\% and 99.73\% at the 256 and 512 resolutions. In the last two columns, we also presented the bit-wise watermark recovery rates of FractalForensics as references, where reasonably higher values than the patch-wise ones can be observed as expected since the patch-wise watermark recovery rate is computed following a more strict rule. 

When tested against Deepfake manipulations, the watermark fragility is expected to be observed. In other words, there is expected to be an obvious gap between the watermark recovery rates for robustness and fragility, such that real and fake samples are distinguishable accordingly. For this experiment, we adopted five face swapping~\cite{SimSwap[15],InfoSwap[5],UniFace[6],E4S[7],DiffSwap[8]} and three face reenactment~\cite{StarGAN-V2[9],StyleMask[10],HyperReenact[11]} manipulations for evaluation. It can be observed from Table~\ref{tab:fake_recovery_rate} that the fragility has been achieved by WaterLo~\cite{WaterLo[1]}, FaceSigns~\cite{FaceSigns[3]}, EditGuard~\cite{EditGuard[4]}, and our proposed FractalForensics, while SepMark~\cite{SepMark[2]} demonstrates robust watermark recovery ability from images generated by SimSwap~\cite{SimSwap[15]}, InfoSwap~\cite{InfoSwap[5]}, and E4S~\cite{E4S[7]}. Meanwhile, FractalForensics is observed to demonstrate promising fragility at both resolutions with average watermark recovery rates of 50.05\% and 39.27\%.

With a closer look at Table~\ref{tab:common_recovery_rate} and Table~\ref{tab:fake_recovery_rate}, since there exist watermark recovery rates against common image operations that are lower than those against Deepfake manipulations for the comparative models (e.g., Sepmark has 99.94\% for InfoSwap but 54.47\% for Gaussian Noise), it becomes impossible to distinguish real and fake based on robustness and fragility. As a result, although our FractalForensics demonstrates slightly higher watermark recovery rates than the other semi-fragile watermarks when tested against face swapping algorithms, there remains a clear gap in the recovery rates regarding robustness and fragility. Additionally, since our watermark matrices are embedded following an entry-to-patch strategy for localizing Deepfake manipulations, only the edited patches are expected to lose the embedded watermarks. Therefore, the critically lower watermark recovery rates for face reenactment than those for face swapping are reasonable and consistent with the localization results in Section~\ref{sec:localization}.

\subsection{Deepfake Detection}

Considering the robustness and fragility of watermarks regarding real and Deepfake images, semi-fragile watermarks act as a natural proactive Deepfake detector. In particular, an ideal case is expected to showcase watermark recovery rates of Deepfake images all lower than those of real images, demonstrating a clear threshold between real and fake samples. In this study, following this scheme, we evaluated the Deepfake detection performance of semi-fragile watermarking frameworks. Since no clear threshold has been provided for the comparative models, we adopted the AUC scores for evaluation. In addition, we included four state-of-the-art passive Deepfake detectors~\cite{FFPP[16],SBIs[17],RECCE[18],CADDM[19]} that have shown reasonable performance on lab-controlled datasets for comparison. In this experiment, the real samples are derived by applying common image operations as listed in Table~\ref{tab:common_recovery_rate} and the fake ones are produced by performing Deepfake manipulations on the real images in CelebA-HQ. 

\begin{figure}
\centering
\includegraphics[width=\columnwidth]{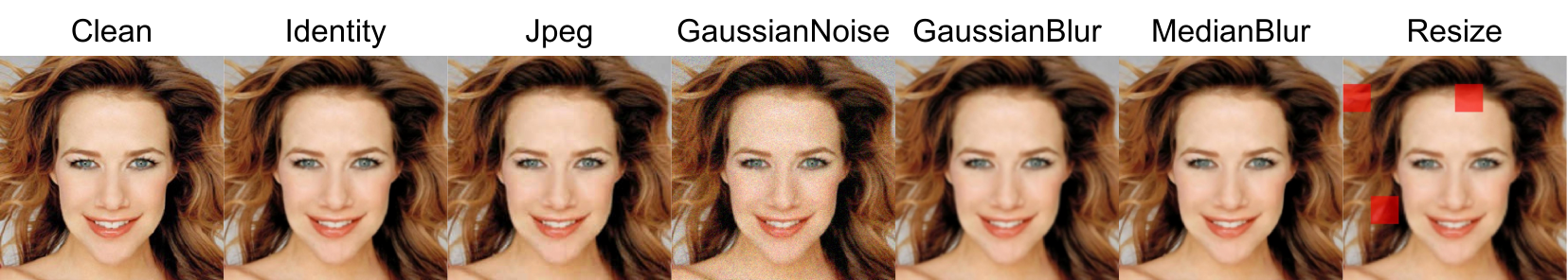}
\caption{Effect of image operations on the watermarked image with localization demonstrated by the red overlay. }
\label{fig:common_localization}
\end{figure}

As Table~\ref{tab:deepfake_detection} lists, although demonstrating some acceptable detection results (e.g., SBIs~\cite{SBIs[17]} and CADDM~\cite{CADDM[19]} on SimSwap), the passive detectors generally exhibit poor generalizability toward most Deepfake manipulations in the experiments, leading to average AUC scores below 70\%. On the other hand, while most semi-fragile algorithms have exhibited desired watermark fragility against most Deepfake manipulations, the proactive Deepfake detection performance depends on their robustness against common image operations. As a result, since a clear gap between watermark recovery rates for real and fake can be observed for FractalForensics, we promisingly achieved satisfactory Deepfake detection results with an average of 99.99\% AUC score over all participating Deepfake manipulations. As for the comparative ones, due to larger watermark recovery rates regarding fragility than those for robustness, WaterLo~\cite{WaterLo[1]} and EditGuard~\cite{EditGuard[4]} fail to achieve Deepfake detection with poor AUC scores. SepMark~\cite{SepMark[2]}, although shows satisfactory performance with AUC scores of 99.99\% for some Deepfake models, the surprisingly high watermark recovery rates in Table~\ref{tab:fake_recovery_rate} result in much lower AUC scores for InfoSwap~\cite{InfoSwap[5]} and E4S~\cite{E4S[7]}, leading to an average of 92.23\% in Table~\ref{tab:deepfake_detection}. Meanwhile, although the watermark recovery rates are close for robustness and fragility, there exists a small gap that benefits FaceSigns~\cite{FaceSigns[3]} in deriving reasonable AUC scores against all Deepfake manipulations. However, its fragility against Gaussian Noise remains a potential risk.

\begin{figure}
\centering
\includegraphics[width=\columnwidth]{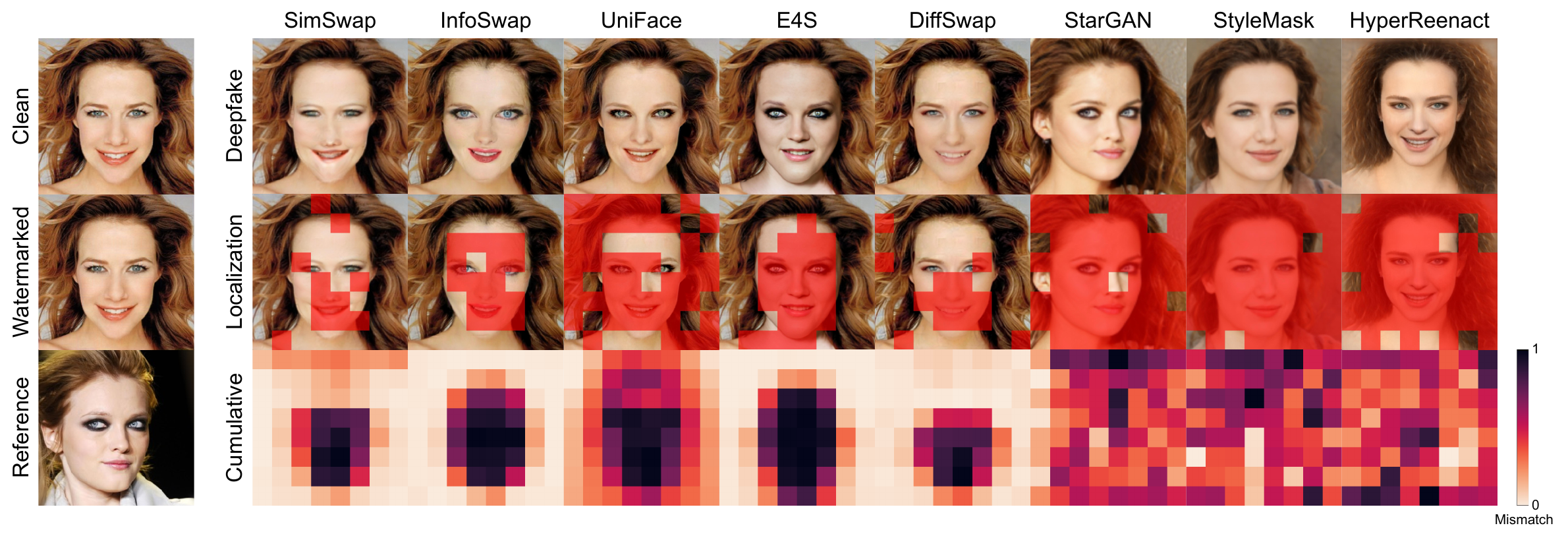}
\caption{Effect of Deepfake algorithms on the watermarked image with localization demonstrated by the red overlay.}
\label{fig:fake_localization}
\end{figure}

\subsection{Deepfake Manipulation Localization}
\label{sec:localization}

\noindent\textbf{Localization Against Common Image Operations.} We first visualized the localization results of our approach when applying common image operations to watermarked images. Since the watermarks are expected to demonstrate robustness when facing common image operations that imitate real-life cases, it is considered a good performance if localized areas barely appear. As Figure~\ref{fig:common_localization} demonstrates, patches that lose watermarks are highlighted with red overlays. In particular,
the exhibition is convincingly consistent with the statistics reported in Table~\ref{tab:common_recovery_rate}, such that FractalForensics is observed to be more sensitive to the Resize operation. In summary, the behavior of the watermarks matches the expectation. 



\noindent\textbf{Localization Against Deepfake Manipulations.} We then evaluated the localization results against Deepfake manipulations. Since Deepfake images are produced upon feature fusion via deep neural networks, there is no ground-truth for localizing Deepfake manipulations. Therefore, FractalForensics is designed to embed watermarks with an entry-to-patch strategy that implicitly enables the localization functionality. As a result, as visualized in Figure~\ref{fig:fake_localization}, the first and second rows display the clean Deepfake results via different manipulation algorithms and their corresponding localization results regarding the sample pair of images, respectively. In the third row, we recorded the number of 4-bit watermark entries that were lost for each patch position on the entire testing dataset and plotted the cumulative localization results after normalization. Given the fact that CelebA-HQ contains facial images that are well-aligned with faces mostly located in the center of images, it can be concluded that, since face swapping manipulations mostly modify detailed facial organs and cues, the localized manipulated areas are mainly in the center of the images. It is also noticed that the localized areas for DiffSwap~\cite{DiffSwap[8]} are mainly in the bottom half of the face. This is because DiffSwap does not heavily edit the forehead and eyes but the nose and mouth. On the other hand, face reenactment algorithms edit not only facial expressions that are located within the face area but also the head poses that can heavily affect the background area, leading to localized areas that largely span entire images. Additionally, by taking a look at Table~\ref{tab:fake_recovery_rate} together with Figure~\ref{fig:fake_localization}, consistencies can be observed between localization visualizations and watermark recovery rates.

\begin{figure}[t!]
\centering
\includegraphics[width=\columnwidth]{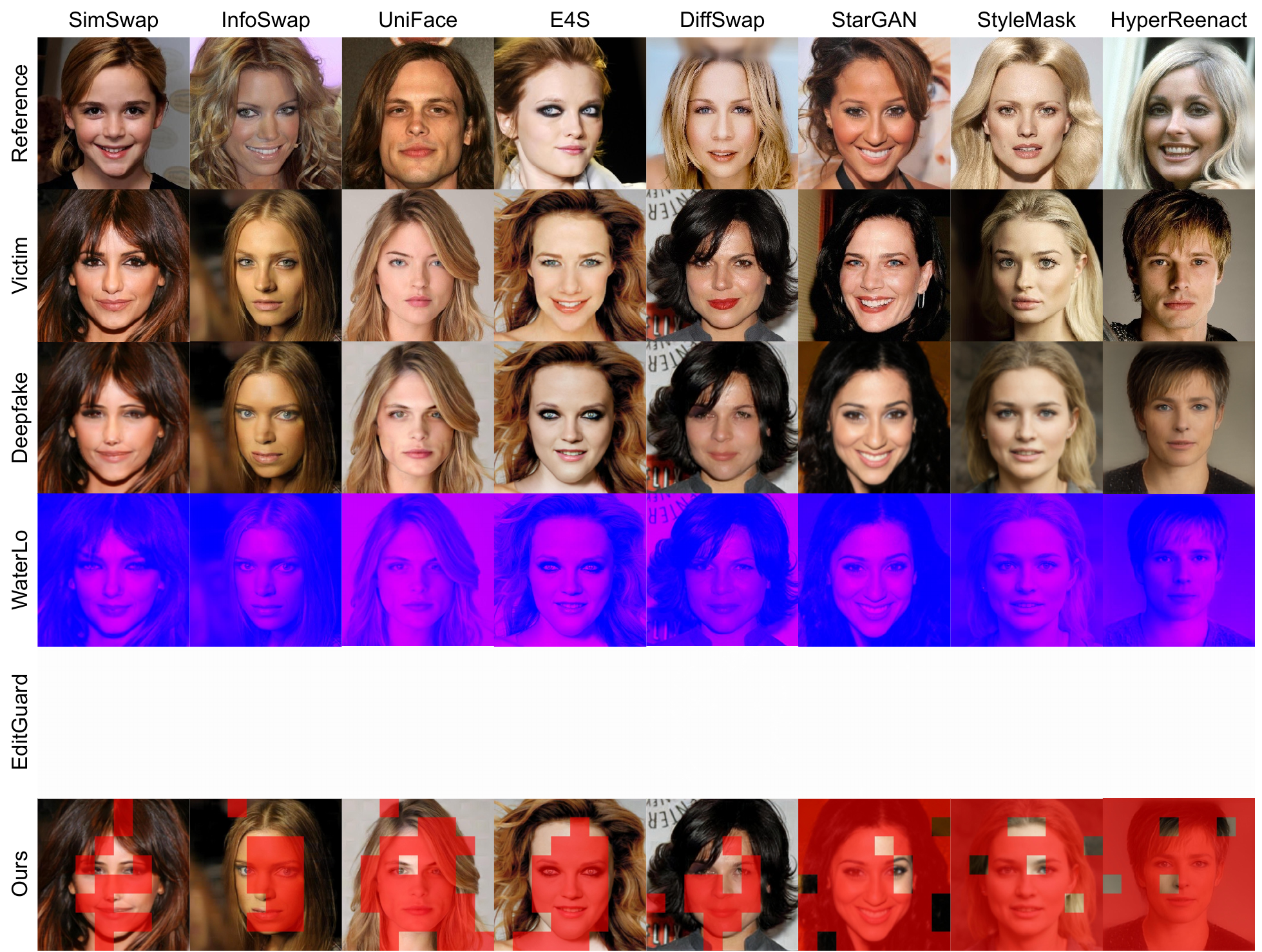}
\caption{Qualitative comparison of localization performance between watermarking frameworks. The localized fake areas are highlighted in purple, white, and red by WaterLo, EditGuard, and our proposed FractalForensics, respectively. }
\label{fig:local_compare}
\end{figure}

\begin{table*}[t!]
\centering
\caption{Cross-dataset evaluation on LFW for visual quality and watermark recovery rate against common image operations.}
\resizebox{\textwidth}{!}{
\begin{tabular}{lcccccc|c|ccc}
\toprule
Models & Identity & Jpeg & Gaussian Noise & Gaussian Blur & Median Blur & Resize & Average$\uparrow$ & PSNR$\uparrow$ & SSIM$\uparrow$ & LPIPS$\downarrow$ \\
\midrule
WaterLo~\cite{WaterLo[1]} & 99.89\% & 61.84\% & 64.53\% & 72.96\% & 99.59\% & 61.87\% & 76.78\% & \textbf{46.807} & \textbf{0.991} & \textbf{0.001} \\
SepMark~\cite{SepMark[2]} & 99.99\% & 99.78\% & 88.05\% & 99.99\% & 99.99\% & 99.99\% & \underline{97.97\%} & 30.825 & 0.947 & 0.021 \\
FaceSigns~\cite{FaceSigns[3]} & 99.98\% & 99.24\% & 66.61\% & 99.95\% & 99.91\% & 99.43\% & 94.19\% & 32.251 & 0.955 & 0.026 \\
EditGuard~\cite{EditGuard[4]} & 99.95\% & 78.23\% & 58.54\% & 0.10\% & 0.81\% & 1.93\% & 39.92\% & \underline{37.880} & 0.749 & 0.031 \\
FractalForensics & 99.95\% & 99.53\% & 99.35\% & 98.97\% & 99.50\% & 97.16\% & \textbf{99.08\%} & 35.741 & \underline{0.970} & \underline{0.019} \\
\bottomrule
\end{tabular}
}
\label{tab:lfw_common}
\end{table*}

\begin{table*}
\centering
\caption{Cross-dataset evaluation on LFW for watermark recovery rate against Deepfake manipulations. }
\resizebox{\textwidth}{!}{
\begin{tabular}{lcccccccc|c}
\toprule
Model & SimSwap~\cite{SimSwap[15]} & InfoSwap~\cite{InfoSwap[5]} & UniFace~\cite{UniFace[6]} & E4S~\cite{E4S[7]} & DiffSwap~\cite{DiffSwap[8]} & StarGAN~\cite{StarGAN-V2[9]} & StyleMask~\cite{StyleMask[10]} & HyperReenact~\cite{HyperReenact[11]} & Average$\downarrow$ \\
\midrule
WaterLo~\cite{WaterLo[1]} & 62.07\% & 45.63\% & 61.84\% & 62.42\% & 62.07\% & 62.14\% & 65.93\% & 62.81\% & 60.74\% \\
SepMark~\cite{SepMark[2]} & 69.57\% & 98.14\% & 52.39\% & 99.87\% & 50.16\% & 50.00\% & 49.68\% & 50.02\% & 64.31\% \\
FaceSigns~\cite{FaceSigns[3]} & 50.08\% & 50.31\% & 49.89\% & 50.03\% & 50.71\% & 50.03\% & 49.92\% & 50.28\% & \underline{56.97\%} \\
EditGuard~\cite{EditGuard[4]} & 49.35\% & 42.48\% & 49.16\% & 50.35\% & 50.49\% & 50.05\% & 49.94\% & 49.99\% & 65.30\% \\
FractalForensics & 74.87\% & 68.57\% & 51.49\% & 82.60\% & 91.48\% & 9.97\% & 10.00\% & 10.08\% & \textbf{49.88\%} \\
\bottomrule
\end{tabular}
}
\label{tab:lfw_deepfake}
\end{table*}

Furthermore, as WaterLo~\cite{WaterLo[1]} and EditGuard~\cite{EditGuard[4]} are also designed to have localization functionality against Deepfake and AIGC, we conducted comparative experiments for the models. The localization results are shown in Figure~\ref{fig:local_compare}. Specifically, WaterLo defines purple overlays as fake areas and uses green for real, while EditGuard highlights the fake and real areas with white and black colors, respectively. However, although successfully highlighting the entire image as fake for all Deepfake-manipulated images, they both lack detailed information that highlights specific manipulated areas. On the other hand, since they each are vulnerable to some common image operations, the localization results become relatively unconvincing. In contrast, Figures~\ref{fig:common_localization},~\ref{fig:fake_localization}, and~\ref{fig:local_compare} show the reliability of FractalForensics in localizing Deepfake manipulated areas and provide explainable evidence to support Deepfake detection. 

\begin{figure}[t!]
\centering
\includegraphics[width=\columnwidth]{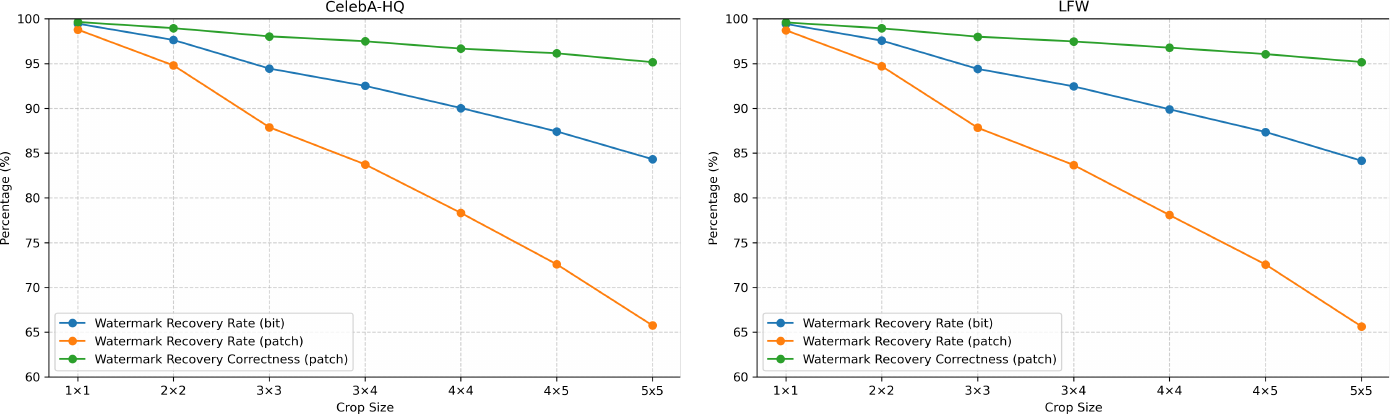}
\caption{Watermark robustness and localization performance against cropping on CelebA-HQ and LFW. }
\label{fig:cropping_robustness}
\end{figure}

\noindent\textbf{Watermark Sensitivity Against Cropping.} In this study, since the watermarking pipeline follows an entry-to-patch strategy, as a side contribution, our approach also exhibits sensitivity against malicious cropping. To explore the watermark robustness in this case, we conducted experiments to randomly crop adjacent image patches and verify whether watermarks in the cropped areas were lost and those in the untampered areas remained. As demonstrated in Figure~\ref{fig:cropping_robustness}, we attempted cropping sizes from $1 \times 1$ to $5 \times 5$ patches, and we recorded the bit-wise and patch-wise watermark recovery rates, along with the correctness regarding desired robustness and fragility. As a result, the bit-wise and patch-wise watermark recovery rates exhibit declines that generally match with the number of cropped patches for both CelebA-HQ and LFW, while the overall correctness stays above 95\% even when 25 out of 64 image patches are cropped within an image. 

\subsection{Cross-Dataset Evaluation on LFW}

In this section, to evaluate the cross-dataset performance of our proposed FractalForensics, we conducted experiments on LFW~\cite{LFW[48]}. Specifically, experiments follow the same settings as in Table~\ref{tab:common_recovery_rate} and Table~\ref{tab:fake_recovery_rate} to compute the watermark recovery rates against common image operations and Deepfake manipulations. For simplicity, the 256 resolution versions for SepMark~\cite{SepMark[2]} and our FractalForensics are utilized in the comparison. 

As Table~\ref{tab:lfw_common} reports, the comparative watermarking frameworks generally maintain consistent performance as for CelebA-HQ~\cite{CelebAHQ[47]}, demonstrating obvious vulnerability when facing Gaussian Noise, with no watermark recovery rate above 90\%. In contrast, although still slightly affected by the Resize operation, our proposed FractalForensics generally preserves robustness against all common image operations at an average of 99.08\% with good visual quality. Additionally, while showing outstanding visual quality, WaterLo fails to achieve the expected watermark robustness. Besides the unsatisfactory robustness and a promising PSNR value, EditGuard exhibits an unexpectedly low SSIM value, which may be caused by mismatched image color or tone. 

Similar outcomes can be seen in Table~\ref{tab:lfw_deepfake} when tested against Deepfake manipulations, such that the comparative models mostly demonstrate fragility while SepMark~\cite{SepMark[2]} is surprisingly robust against InfoSwap~\cite{InfoSwap[5]} and E4S~\cite{E4S[7]}. Consequently, the proposed FractalForensics obtains an optimal average watermark recovery rate of 49.88\% in this comparison. Furthermore, when taking a look at Table~\ref{tab:lfw_common} and Table~\ref{tab:lfw_deepfake} together, a clear gap for watermark robustness and fragility regarding common image operations and Deepfake manipulations can still be found for our proposed FractalForensics. In conclusion, our approach performs satisfactorily even under the cross-dataset setting, outperforming the state-of-the-art semi-fragile watermarking frameworks. 

\section{Conclusion}

In this work, we propose FractalForensics, a novel proactive Deepfake detection method based on semi-fragile fractal watermarks. Relying on the characteristics of fractal shapes, we first design a parameter-driven pipeline for watermark generation, allowing user-selected parameters to determine the variations toward the fractal shape and the watermark encryption rule via the chaotic encryption system. Upon placing the 4-bit watermark matrix entries at the channel dimension, we introduce an innovative entry-to-patch strategy for position-aware watermark embedding and recovery, which implicitly helps accomplish explainable localizations of malicious Deepfakes. In conclusion, FractalForensics achieves excellent robustness and fragility for the watermark recovery rates, leading to promising proactive Deepfake detection and localization performance, outperforming state-of-the-art passive and semi-fragile approaches. Future work may continue to exploit the self-similarity of fractal shapes, where an automatic self-check can be applied to the recovered watermark without knowing the parameters. If capacity allows, an auxiliary robust watermark may also be integrated for additional source-tracing functionality. Moreover, since real-life cases are complex and contain not only facial parts, further developments are still required for actual deployment\footnote{This work has been accepted to ACM Multimedia 2025 as an oral paper. }. 

\begin{acks}
This research is supported by the Ministry of Education, Singapore, under its MOE AcRF TIER 3 Grant (MOE-MOET32022-0001) and by VoiceCode Technology Pte. Ltd.
\end{acks}

\bibliographystyle{ACM-Reference-Format}
\bibliography{sample-base}


\begin{thebibliography}{53}


\ifx \showCODEN    \undefined \def \showCODEN     #1{\unskip}     \fi
\ifx \showISBNx    \undefined \def \showISBNx     #1{\unskip}     \fi
\ifx \showISBNxiii \undefined \def \showISBNxiii  #1{\unskip}     \fi
\ifx \showISSN     \undefined \def \showISSN      #1{\unskip}     \fi
\ifx \showLCCN     \undefined \def \showLCCN      #1{\unskip}     \fi
\ifx \shownote     \undefined \def \shownote      #1{#1}          \fi
\ifx \showarticletitle \undefined \def \showarticletitle #1{#1}   \fi
\ifx \showURL      \undefined \def \showURL       {\relax}        \fi
\providecommand\bibfield[2]{#2}
\providecommand\bibinfo[2]{#2}
\providecommand\natexlab[1]{#1}
\providecommand\showeprint[2][]{arXiv:#2}

\bibitem[Allezaud(1985)]%
        {Allezaud_1985[24]}
\bibfield{author}{\bibinfo{person}{Robert Allezaud}.} \bibinfo{year}{1985}\natexlab{}.
\newblock \showarticletitle{Les objets fractals, par B. Mandelbrot}.
\newblock \bibinfo{journal}{\emph{Communication \& Langages}} \bibinfo{volume}{64}, \bibinfo{number}{1} (\bibinfo{date}{Jan.} \bibinfo{year}{1985}), \bibinfo{pages}{123–124}.
\newblock


\bibitem[Bader(2000)]%
        {ZorderMemoryLayout2000[49]}
\bibfield{author}{\bibinfo{person}{Michael Bader}.} \bibinfo{year}{2000}\natexlab{}.
\newblock \showarticletitle{Improving cache performance with the Z-order curve}. In \bibinfo{booktitle}{\emph{Lecture Notes in Computer Science}}, Vol.~\bibinfo{volume}{1900}. \bibinfo{publisher}{Springer}, \bibinfo{pages}{253--262}.
\newblock


\bibitem[Beuve et~al\mbox{.}(2023)]%
        {WaterLo[1]}
\bibfield{author}{\bibinfo{person}{Nicolas Beuve}, \bibinfo{person}{Wassim Hamidouche}, {and} \bibinfo{person}{Olivier D\'eforges}.} \bibinfo{year}{2023}\natexlab{}.
\newblock \showarticletitle{WaterLo: Protect Images from Deepfakes Using Localized Semi-Fragile Watermark}. In \bibinfo{booktitle}{\emph{Proceedings of the IEEE/CVF International Conference on Computer Vision Workshops}}. \bibinfo{pages}{393--402}.
\newblock


\bibitem[Bounareli et~al\mbox{.}(2023a)]%
        {HyperReenact[11]}
\bibfield{author}{\bibinfo{person}{S. Bounareli}, \bibinfo{person}{C. Tzelepis}, \bibinfo{person}{V. Argyriou}, \bibinfo{person}{I. Patras}, {and} \bibinfo{person}{G. Tzimiropoulos}.} \bibinfo{year}{2023}\natexlab{a}.
\newblock \showarticletitle{HyperReenact: One-Shot Reenactment via Jointly Learning to Refine and Retarget Faces}. In \bibinfo{booktitle}{\emph{Proceedings of the IEEE/CVF International Conference on Computer Vision}}. \bibinfo{pages}{7115--7125}.
\newblock


\bibitem[Bounareli et~al\mbox{.}(2023b)]%
        {StyleMask[10]}
\bibfield{author}{\bibinfo{person}{S. Bounareli}, \bibinfo{person}{C. Tzelepis}, \bibinfo{person}{V. Argyriou}, \bibinfo{person}{I. Patras}, {and} \bibinfo{person}{G. Tzimiropoulos}.} \bibinfo{year}{2023}\natexlab{b}.
\newblock \showarticletitle{StyleMask: Disentangling the Style Space of StyleGAN2 for Neural Face Reenactment}. In \bibinfo{booktitle}{\emph{IEEE 17th International Conference on Automatic Face and Gesture Recognition}}. \bibinfo{pages}{1--8}.
\newblock


\bibitem[Cao et~al\mbox{.}(2022)]%
        {RECCE[18]}
\bibfield{author}{\bibinfo{person}{Junyi Cao}, \bibinfo{person}{Chao Ma}, \bibinfo{person}{Taiping Yao}, \bibinfo{person}{Shen Chen}, \bibinfo{person}{Shouhong Ding}, {and} \bibinfo{person}{Xiaokang Yang}.} \bibinfo{year}{2022}\natexlab{}.
\newblock \showarticletitle{End-to-End Reconstruction-Classification Learning for Face Forgery Detection}. In \bibinfo{booktitle}{\emph{Proceedings of the IEEE/CVF Conference on Computer Vision and Pattern Recognition}}. \bibinfo{pages}{4113--4122}.
\newblock


\bibitem[Chen et~al\mbox{.}(2020)]%
        {SimSwap[15]}
\bibfield{author}{\bibinfo{person}{Renwang Chen}, \bibinfo{person}{Xuanhong Chen}, \bibinfo{person}{Bingbing Ni}, {and} \bibinfo{person}{Yanhao Ge}.} \bibinfo{year}{2020}\natexlab{}.
\newblock \showarticletitle{SimSwap: An Efficient Framework For High Fidelity Face Swapping}. In \bibinfo{booktitle}{\emph{The 28th ACM International Conference on Multimedia}}. \bibinfo{pages}{2003–2011}.
\newblock
\showISBNx{9781450379885}
\href{https://doi.org/10.1145/3394171.3413630}{doi:\nolinkurl{10.1145/3394171.3413630}}


\bibitem[Choi et~al\mbox{.}(2020)]%
        {StarGAN-V2[9]}
\bibfield{author}{\bibinfo{person}{Yunjey Choi}, \bibinfo{person}{Youngjung Uh}, \bibinfo{person}{Jaejun Yoo}, {and} \bibinfo{person}{Jung-Woo Ha}.} \bibinfo{year}{2020}\natexlab{}.
\newblock \showarticletitle{StarGAN v2: Diverse Image Synthesis for Multiple Domains}. In \bibinfo{booktitle}{\emph{Proceedings of the IEEE/CVF Conference on Computer Vision and Pattern Recognition}}. \bibinfo{pages}{8185--8194}.
\newblock


\bibitem[Dong et~al\mbox{.}(2023)]%
        {CADDM[19]}
\bibfield{author}{\bibinfo{person}{Shichao Dong}, \bibinfo{person}{Jin Wang}, \bibinfo{person}{Renhe Ji}, \bibinfo{person}{Jiajun Liang}, \bibinfo{person}{Haoqiang Fan}, {and} \bibinfo{person}{Zheng Ge}.} \bibinfo{year}{2023}\natexlab{}.
\newblock \showarticletitle{Implicit Identity Leakage: The Stumbling Block to Improving Deepfake Detection Generalization}. In \bibinfo{booktitle}{\emph{Proceedings of the IEEE/CVF Conference on Computer Vision and Pattern Recognition}}. \bibinfo{pages}{3994--4004}.
\newblock


\bibitem[Gao et~al\mbox{.}(2021)]%
        {InfoSwap[5]}
\bibfield{author}{\bibinfo{person}{Gege Gao}, \bibinfo{person}{Huaibo Huang}, \bibinfo{person}{Chaoyou Fu}, \bibinfo{person}{Zhaoyang Li}, {and} \bibinfo{person}{Ran He}.} \bibinfo{year}{2021}\natexlab{}.
\newblock \showarticletitle{Information Bottleneck Disentanglement for Identity Swapping}. In \bibinfo{booktitle}{\emph{Proceedings of the IEEE/CVF Conference on Computer Vision and Pattern Recognition}}. \bibinfo{pages}{3404--3413}.
\newblock


\bibitem[He et~al\mbox{.}(2016)]%
        {ResNet[44]}
\bibfield{author}{\bibinfo{person}{Kaiming He}, \bibinfo{person}{X. Zhang}, \bibinfo{person}{Shaoqing Ren}, {and} \bibinfo{person}{Jian Sun}.} \bibinfo{year}{2016}\natexlab{}.
\newblock \showarticletitle{Deep Residual Learning for Image Recognition}.
\newblock \bibinfo{journal}{\emph{IEEE Conference on Computer Vision and Pattern Recognition}} (\bibinfo{year}{2016}), \bibinfo{pages}{770--778}.
\newblock


\bibitem[Hilbert(1891)]%
        {Hilbert1891[43]}
\bibfield{author}{\bibinfo{person}{David Hilbert}.} \bibinfo{year}{1891}\natexlab{}.
\newblock \showarticletitle{Ueber die stetige Abbildung einer Line auf ein Fl{\"a}chenst{\"u}ck}.
\newblock \bibinfo{journal}{\emph{Math. Ann.}} \bibinfo{volume}{38}, \bibinfo{number}{3} (\bibinfo{date}{01 Sep} \bibinfo{year}{1891}), \bibinfo{pages}{459--460}.
\newblock
\showISSN{1432-1807}
\href{https://doi.org/10.1007/BF01199431}{doi:\nolinkurl{10.1007/BF01199431}}


\bibitem[Hu et~al\mbox{.}(2018)]%
        {SENet[45]}
\bibfield{author}{\bibinfo{person}{Jie Hu}, \bibinfo{person}{Li Shen}, {and} \bibinfo{person}{Gang Sun}.} \bibinfo{year}{2018}\natexlab{}.
\newblock \showarticletitle{Squeeze-and-Excitation Networks}. In \bibinfo{booktitle}{\emph{IEEE Conference on Computer Vision and Pattern Recognition}}. \bibinfo{pages}{7132--7141}.
\newblock
\href{https://doi.org/10.1109/CVPR.2018.00745}{doi:\nolinkurl{10.1109/CVPR.2018.00745}}


\bibitem[Huang et~al\mbox{.}(2012)]%
        {LFW[48]}
\bibfield{author}{\bibinfo{person}{Gary~B. Huang}, \bibinfo{person}{Marwan Mattar}, \bibinfo{person}{Honglak Lee}, {and} \bibinfo{person}{Erik Learned-Miller}.} \bibinfo{year}{2012}\natexlab{}.
\newblock \showarticletitle{Learning to Align from Scratch}. In \bibinfo{booktitle}{\emph{Advances in Neural Information Processing Systems}}.
\newblock


\bibitem[Huang et~al\mbox{.}(2022)]%
        {CMUA[28]}
\bibfield{author}{\bibinfo{person}{Hao Huang}, \bibinfo{person}{Yongtao Wang}, \bibinfo{person}{Zhaoyu Chen}, \bibinfo{person}{Yuze Zhang}, \bibinfo{person}{Yuheng Li}, \bibinfo{person}{Zhi Tang}, \bibinfo{person}{Wei Chu}, \bibinfo{person}{Jingdong Chen}, \bibinfo{person}{Weisi Lin}, {and} \bibinfo{person}{Kai-Kuang Ma}.} \bibinfo{year}{2022}\natexlab{}.
\newblock \showarticletitle{CMUA-Watermark: A Cross-Model Universal Adversarial Watermark for Combating Deepfakes}.
\newblock \bibinfo{journal}{\emph{Proceedings of the AAAI Conference on Artificial Intelligence}} \bibinfo{volume}{36}, \bibinfo{number}{1} (\bibinfo{date}{6} \bibinfo{year}{2022}), \bibinfo{pages}{989--997}.
\newblock
\href{https://doi.org/10.1609/aaai.v36i1.19982}{doi:\nolinkurl{10.1609/aaai.v36i1.19982}}


\bibitem[Jia et~al\mbox{.}(2021)]%
        {MBRS[32]}
\bibfield{author}{\bibinfo{person}{Zhaoyang Jia}, \bibinfo{person}{Han Fang}, {and} \bibinfo{person}{Weiming Zhang}.} \bibinfo{year}{2021}\natexlab{}.
\newblock \showarticletitle{MBRS: Enhancing Robustness of DNN-based Watermarking by Mini-Batch of Real and Simulated JPEG Compression}. In \bibinfo{booktitle}{\emph{Proceedings of the 29th ACM International Conference on Multimedia}}. \bibinfo{pages}{41–49}.
\newblock
\showISBNx{9781450386517}


\bibitem[Karras et~al\mbox{.}(2018)]%
        {CelebAHQ[47]}
\bibfield{author}{\bibinfo{person}{Tero Karras}, \bibinfo{person}{Timo Aila}, \bibinfo{person}{Samuli Laine}, {and} \bibinfo{person}{Jaakko Lehtinen}.} \bibinfo{year}{2018}\natexlab{}.
\newblock \showarticletitle{Progressive Growing of {GAN}s for Improved Quality, Stability, and Variation}. In \bibinfo{booktitle}{\emph{International Conference on Learning Representations}}.
\newblock


\bibitem[Kong et~al\mbox{.}(2022)]%
        {kong2022detect[52]}
\bibfield{author}{\bibinfo{person}{Chenqi Kong}, \bibinfo{person}{Baoliang Chen}, \bibinfo{person}{Haoliang Li}, \bibinfo{person}{Shiqi Wang}, \bibinfo{person}{Anderson Rocha}, {and} \bibinfo{person}{Sam Kwong}.} \bibinfo{year}{2022}\natexlab{}.
\newblock \showarticletitle{Detect and locate: Exposing face manipulation by semantic-and noise-level telltales}.
\newblock \bibinfo{journal}{\emph{IEEE Transactions on Information Forensics and Security}}  \bibinfo{volume}{17} (\bibinfo{year}{2022}), \bibinfo{pages}{1741--1756}.
\newblock


\bibitem[Kong et~al\mbox{.}(2024)]%
        {kong2024moe[54]}
\bibfield{author}{\bibinfo{person}{Chenqi Kong}, \bibinfo{person}{Anwei Luo}, \bibinfo{person}{Peijun Bao}, \bibinfo{person}{Yi Yu}, \bibinfo{person}{Haoliang Li}, \bibinfo{person}{Zengwei Zheng}, \bibinfo{person}{Shiqi Wang}, {and} \bibinfo{person}{Alex~C Kot}.} \bibinfo{year}{2024}\natexlab{}.
\newblock \showarticletitle{Moe-ffd: Mixture of experts for generalized and parameter-efficient face forgery detection}.
\newblock \bibinfo{journal}{\emph{arXiv preprint arXiv:2404.08452}} (\bibinfo{year}{2024}).
\newblock


\bibitem[Kong et~al\mbox{.}(2025)]%
        {kong2025pixel[53]}
\bibfield{author}{\bibinfo{person}{Chenqi Kong}, \bibinfo{person}{Anwei Luo}, \bibinfo{person}{Shiqi Wang}, \bibinfo{person}{Haoliang Li}, \bibinfo{person}{Anderson Rocha}, {and} \bibinfo{person}{Alex~C Kot}.} \bibinfo{year}{2025}\natexlab{}.
\newblock \showarticletitle{Pixel-inconsistency modeling for image manipulation localization}.
\newblock \bibinfo{journal}{\emph{IEEE Transactions on Pattern Analysis and Machine Intelligence}} (\bibinfo{year}{2025}).
\newblock


\bibitem[Krizhevsky et~al\mbox{.}(2012)]%
        {AlexNet[46]}
\bibfield{author}{\bibinfo{person}{Alex Krizhevsky}, \bibinfo{person}{Ilya Sutskever}, {and} \bibinfo{person}{Geoffrey~E. Hinton}.} \bibinfo{year}{2012}\natexlab{}.
\newblock \showarticletitle{ImageNet classification with deep convolutional neural networks}.
\newblock \bibinfo{journal}{\emph{Commun. ACM}}  \bibinfo{volume}{60} (\bibinfo{year}{2012}), \bibinfo{pages}{84 -- 90}.
\newblock


\bibitem[Liu et~al\mbox{.}(2023)]%
        {E4S[7]}
\bibfield{author}{\bibinfo{person}{Zhian Liu}, \bibinfo{person}{Maomao Li}, \bibinfo{person}{Yong Zhang}, \bibinfo{person}{Cairong Wang}, \bibinfo{person}{Qi Zhang}, \bibinfo{person}{Jue Wang}, {and} \bibinfo{person}{Yongwei Nie}.} \bibinfo{year}{2023}\natexlab{}.
\newblock \showarticletitle{Fine-Grained Face Swapping via Regional GAN Inversion}. In \bibinfo{booktitle}{\emph{Proceedings of the IEEE/CVF Conference on Computer Vision and Pattern Recognition}}. \bibinfo{pages}{8578--8587}.
\newblock


\bibitem[Luo et~al\mbox{.}(2024)]%
        {luo2024forgery[55]}
\bibfield{author}{\bibinfo{person}{Anwei Luo}, \bibinfo{person}{Rizhao Cai}, \bibinfo{person}{Chenqi Kong}, \bibinfo{person}{Yakun Ju}, \bibinfo{person}{Xiangui Kang}, \bibinfo{person}{Jiwu Huang}, {and} \bibinfo{person}{Alex C~Kot Life}.} \bibinfo{year}{2024}\natexlab{}.
\newblock \showarticletitle{Forgery-aware adaptive learning with vision transformer for generalized face forgery detection}.
\newblock \bibinfo{journal}{\emph{IEEE Transactions on Circuits and Systems for Video Technology}} (\bibinfo{year}{2024}).
\newblock


\bibitem[Luo et~al\mbox{.}(2023)]%
        {luo2023beyond[56]}
\bibfield{author}{\bibinfo{person}{Anwei Luo}, \bibinfo{person}{Chenqi Kong}, \bibinfo{person}{Jiwu Huang}, \bibinfo{person}{Yongjian Hu}, \bibinfo{person}{Xiangui Kang}, {and} \bibinfo{person}{Alex~C Kot}.} \bibinfo{year}{2023}\natexlab{}.
\newblock \showarticletitle{Beyond the prior forgery knowledge: Mining critical clues for general face forgery detection}.
\newblock \bibinfo{journal}{\emph{IEEE Transactions on Information Forensics and Security}}  \bibinfo{volume}{19} (\bibinfo{year}{2023}), \bibinfo{pages}{1168--1182}.
\newblock


\bibitem[Ma et~al\mbox{.}(2022)]%
        {CIN[34]}
\bibfield{author}{\bibinfo{person}{Rui Ma}, \bibinfo{person}{Mengxi Guo}, \bibinfo{person}{Yi Hou}, \bibinfo{person}{Fan Yang}, \bibinfo{person}{Yuan Li}, \bibinfo{person}{Huizhu Jia}, {and} \bibinfo{person}{Xiaodong Xie}.} \bibinfo{year}{2022}\natexlab{}.
\newblock \showarticletitle{Towards Blind Watermarking: Combining Invertible and Non-invertible Mechanisms}. In \bibinfo{booktitle}{\emph{Proceedings of the 30th ACM International Conference on Multimedia}}. \bibinfo{pages}{1532–1542}.
\newblock
\showISBNx{9781450392037}


\bibitem[Mandelbrot(1982)]%
        {SaganFractals1994[40]}
\bibfield{author}{\bibinfo{person}{Benoît~B. Mandelbrot}.} \bibinfo{year}{1982}\natexlab{}.
\newblock \bibinfo{booktitle}{\emph{The Fractal Geometry of Nature}}.
\newblock \bibinfo{publisher}{W.H. Freeman and Company}, \bibinfo{address}{New York}.
\newblock
\showISBNx{9780716711865}


\bibitem[Matthews(1989)]%
        {ChaoticEnc[25]}
\bibfield{author}{\bibinfo{person}{Robert Matthews}.} \bibinfo{year}{1989}\natexlab{}.
\newblock \showarticletitle{On the derivation of a “Chaotic” encryption algorithm}.
\newblock \bibinfo{journal}{\emph{Cryptologia}} \bibinfo{volume}{13}, \bibinfo{number}{1} (\bibinfo{year}{1989}), \bibinfo{pages}{29--42}.
\newblock
\href{https://doi.org/10.1080/0161-118991863745}{doi:\nolinkurl{10.1080/0161-118991863745}}


\bibitem[Neekhara et~al\mbox{.}(2024)]%
        {FaceSigns[3]}
\bibfield{author}{\bibinfo{person}{Paarth Neekhara}, \bibinfo{person}{Shehzeen Hussain}, \bibinfo{person}{Xinqiao Zhang}, \bibinfo{person}{Ke Huang}, \bibinfo{person}{Julian McAuley}, {and} \bibinfo{person}{Farinaz Koushanfar}.} \bibinfo{year}{2024}\natexlab{}.
\newblock \showarticletitle{FaceSigns: Semi-fragile Watermarks for Media Authentication}.
\newblock \bibinfo{journal}{\emph{ACM Trans. Multimedia Comput. Commun. Appl.}} \bibinfo{volume}{20}, \bibinfo{number}{11}, Article \bibinfo{articleno}{337} (\bibinfo{date}{Sept.} \bibinfo{year}{2024}), \bibinfo{numpages}{21}~pages.
\newblock
\showISSN{1551-6857}
\href{https://doi.org/10.1145/3640466}{doi:\nolinkurl{10.1145/3640466}}


\bibitem[Qu et~al\mbox{.}(2024)]%
        {DFRAP[29]}
\bibfield{author}{\bibinfo{person}{Zuomin Qu}, \bibinfo{person}{Zuping Xi}, \bibinfo{person}{Wei Lu}, \bibinfo{person}{Xiangyang Luo}, \bibinfo{person}{Qian Wang}, {and} \bibinfo{person}{Bin Li}.} \bibinfo{year}{2024}\natexlab{}.
\newblock \showarticletitle{DF-RAP: A Robust Adversarial Perturbation for Defending against Deepfakes in Real-world Social Network Scenarios}.
\newblock \bibinfo{journal}{\emph{IEEE Transactions on Information Forensics and Security}} (\bibinfo{year}{2024}).
\newblock


\bibitem[Reza and Flickner(1999)]%
        {HilbertImageCompression1999[41]}
\bibfield{author}{\bibinfo{person}{Ali~M. Reza} {and} \bibinfo{person}{Myron~D. Flickner}.} \bibinfo{year}{1999}\natexlab{}.
\newblock \showarticletitle{Image compression using Hilbert space-filling curves}.
\newblock \bibinfo{journal}{\emph{Proceedings of the IEEE International Conference on Image Processing (ICIP)}} (\bibinfo{year}{1999}), \bibinfo{pages}{792--795}.
\newblock
\href{https://doi.org/10.1109/ICIP.1999.819848}{doi:\nolinkurl{10.1109/ICIP.1999.819848}}


\bibitem[Rössler et~al\mbox{.}(2019)]%
        {FFPP[16]}
\bibfield{author}{\bibinfo{person}{Andreas Rössler}, \bibinfo{person}{Davide Cozzolino}, \bibinfo{person}{Luisa Verdoliva}, \bibinfo{person}{Christian Riess}, \bibinfo{person}{Justus Thies}, {and} \bibinfo{person}{Matthias Niessner}.} \bibinfo{year}{2019}\natexlab{}.
\newblock \showarticletitle{FaceForensics++: Learning to Detect Manipulated Facial Images}. In \bibinfo{booktitle}{\emph{Proceedings of the IEEE/CVF International Conference on Computer Vision}}. \bibinfo{pages}{1--11}.
\newblock


\bibitem[Shiohara and Yamasaki(2022)]%
        {SBIs[17]}
\bibfield{author}{\bibinfo{person}{Kaede Shiohara} {and} \bibinfo{person}{Toshihiko Yamasaki}.} \bibinfo{year}{2022}\natexlab{}.
\newblock \showarticletitle{Detecting Deepfakes With Self-Blended Images}. In \bibinfo{booktitle}{\emph{Proceedings of the IEEE/CVF Conference on Computer Vision and Pattern Recognition}}. \bibinfo{pages}{18720--18729}.
\newblock


\bibitem[Sun et~al\mbox{.}(2022)]%
        {sun2022faketracer[39]}
\bibfield{author}{\bibinfo{person}{Pu Sun}, \bibinfo{person}{Yuezun Li}, \bibinfo{person}{Honggang Qi}, {and} \bibinfo{person}{Siwei Lyu}.} \bibinfo{year}{2022}\natexlab{}.
\newblock \showarticletitle{Faketracer: Exposing Deepfakes with Training Data Contamination}. In \bibinfo{booktitle}{\emph{2022 IEEE International Conference on Image Processing}}. \bibinfo{pages}{1161--1165}.
\newblock
\href{https://doi.org/10.1109/ICIP46576.2022.9897756}{doi:\nolinkurl{10.1109/ICIP46576.2022.9897756}}


\bibitem[Sun et~al\mbox{.}(2024)]%
        {sun2024faketracer[38]}
\bibfield{author}{\bibinfo{person}{Pu Sun}, \bibinfo{person}{Honggang Qi}, \bibinfo{person}{Yuezun Li}, {and} \bibinfo{person}{Siwei Lyu}.} \bibinfo{year}{2024}\natexlab{}.
\newblock \bibinfo{title}{FakeTracer: Catching Face-swap DeepFakes via Implanting Traces in Training}.
\newblock


\bibitem[Tan et~al\mbox{.}(2024)]%
        {tan2024rle[51]}
\bibfield{author}{\bibinfo{person}{Lei Tan}, \bibinfo{person}{Yukang Zhang}, \bibinfo{person}{Keke Han}, \bibinfo{person}{Pingyang Dai}, \bibinfo{person}{Yan Zhang}, \bibinfo{person}{Yongjian Wu}, {and} \bibinfo{person}{Rongrong Ji}.} \bibinfo{year}{2024}\natexlab{}.
\newblock \showarticletitle{RLE: A unified perspective of data augmentation for cross-spectral re-identification}.
\newblock \bibinfo{journal}{\emph{Advances in Neural Information Processing Systems}}  \bibinfo{volume}{37} (\bibinfo{year}{2024}), \bibinfo{pages}{126977--126996}.
\newblock


\bibitem[van Leeuwen(1991)]%
        {NP-Hard[50]}
\bibfield{editor}{\bibinfo{person}{Jan van Leeuwen}} (Ed.). \bibinfo{year}{1991}\natexlab{}.
\newblock \bibinfo{booktitle}{\emph{Handbook of theoretical computer science (vol. A): algorithms and complexity}}.
\newblock
\showISBNx{0444880712}


\bibitem[Wang et~al\mbox{.}(2021)]%
        {Wang2021FakeTagger[31]}
\bibfield{author}{\bibinfo{person}{Run Wang}, \bibinfo{person}{Felix Juefei-Xu}, \bibinfo{person}{Meng Luo}, \bibinfo{person}{Yang Liu}, {and} \bibinfo{person}{Lina Wang}.} \bibinfo{year}{2021}\natexlab{}.
\newblock \showarticletitle{FakeTagger: Robust Safeguards against DeepFake Dissemination via Provenance Tracking}. In \bibinfo{booktitle}{\emph{Proceedings of the 29th ACM International Conference on Multimedia}}. \bibinfo{pages}{3546–3555}.
\newblock
\showISBNx{9781450386517}
\href{https://doi.org/10.1145/3474085.3475518}{doi:\nolinkurl{10.1145/3474085.3475518}}


\bibitem[Wang et~al\mbox{.}(2023)]%
        {DCPT[21]}
\bibfield{author}{\bibinfo{person}{Tianyi Wang}, \bibinfo{person}{Harry Cheng}, \bibinfo{person}{Kam~Pui Chow}, {and} \bibinfo{person}{Liqiang Nie}.} \bibinfo{year}{2023}\natexlab{}.
\newblock \showarticletitle{Deep Convolutional Pooling Transformer for Deepfake Detection}.
\newblock \bibinfo{journal}{\emph{ACM Transactions on Multimedia Computing, Communications, and Applications}} \bibinfo{volume}{19}, \bibinfo{number}{6}, Article \bibinfo{articleno}{179} (\bibinfo{year}{2023}), \bibinfo{numpages}{20}~pages.
\newblock
\showISSN{1551-6857}
\href{https://doi.org/10.1145/3588574}{doi:\nolinkurl{10.1145/3588574}}


\bibitem[Wang and Chow(2023)]%
        {NoiseDF[22]}
\bibfield{author}{\bibinfo{person}{Tianyi Wang} {and} \bibinfo{person}{Kam~Pui Chow}.} \bibinfo{year}{2023}\natexlab{}.
\newblock \showarticletitle{Noise Based Deepfake Detection via Multi-Head Relative-Interaction}.
\newblock \bibinfo{journal}{\emph{Proceedings of the AAAI Conference on Artificial Intelligence}} \bibinfo{volume}{37}, \bibinfo{number}{12} (\bibinfo{year}{2023}), \bibinfo{pages}{14548--14556}.
\newblock
\href{https://doi.org/10.1609/aaai.v37i12.26701}{doi:\nolinkurl{10.1609/aaai.v37i12.26701}}


\bibitem[Wang et~al\mbox{.}(2024a)]%
        {IDPMark2024Wang[13]}
\bibfield{author}{\bibinfo{person}{Tianyi Wang}, \bibinfo{person}{Mengxiao Huang}, \bibinfo{person}{Harry Cheng}, \bibinfo{person}{Bin Ma}, {and} \bibinfo{person}{Yinglong Wang}.} \bibinfo{year}{2024}\natexlab{a}.
\newblock \showarticletitle{Robust Identity Perceptual Watermark Against Deepfake Face Swapping}.
\newblock \bibinfo{journal}{\emph{arXiv preprint arXiv:2311.01357}}.
\newblock


\bibitem[Wang et~al\mbox{.}(2024b)]%
        {LampMark2024Wang[12]}
\bibfield{author}{\bibinfo{person}{Tianyi Wang}, \bibinfo{person}{Mengxiao Huang}, \bibinfo{person}{Harry Cheng}, \bibinfo{person}{Xiao Zhang}, {and} \bibinfo{person}{Zhiqi Shen}.} \bibinfo{year}{2024}\natexlab{b}.
\newblock \showarticletitle{LampMark: Proactive Deepfake Detection via Training-Free Landmark Perceptual Watermarks}. In \bibinfo{booktitle}{\emph{Proceedings of the 32nd ACM International Conference on Multimedia}} \emph{(\bibinfo{series}{MM '24})}. \bibinfo{pages}{10515–10524}.
\newblock
\showISBNx{9798400706868}
\href{https://doi.org/10.1145/3664647.3680869}{doi:\nolinkurl{10.1145/3664647.3680869}}


\bibitem[Wang et~al\mbox{.}(2024c)]%
        {APSwap[26]}
\bibfield{author}{\bibinfo{person}{Tianyi Wang}, \bibinfo{person}{Zian Li}, \bibinfo{person}{Ruixia Liu}, \bibinfo{person}{Yinglong Wang}, {and} \bibinfo{person}{Liqiang Nie}.} \bibinfo{year}{2024}\natexlab{c}.
\newblock \showarticletitle{An Efficient Attribute-Preserving Framework for Face Swapping}.
\newblock \bibinfo{journal}{\emph{IEEE Transactions on Multimedia}}  \bibinfo{volume}{26} (\bibinfo{year}{2024}), \bibinfo{pages}{6554--6565}.
\newblock
\href{https://doi.org/10.1109/TMM.2024.3354573}{doi:\nolinkurl{10.1109/TMM.2024.3354573}}


\bibitem[Wang et~al\mbox{.}(2024d)]%
        {DeepfakeSurvey[20]}
\bibfield{author}{\bibinfo{person}{Tianyi Wang}, \bibinfo{person}{Xin Liao}, \bibinfo{person}{Kam~Pui Chow}, \bibinfo{person}{Xiaodong Lin}, {and} \bibinfo{person}{Yinglong Wang}.} \bibinfo{year}{2024}\natexlab{d}.
\newblock \showarticletitle{Deepfake Detection: A Comprehensive Survey from the Reliability Perspective}.
\newblock \bibinfo{journal}{\emph{Comput. Surveys}} (\bibinfo{year}{2024}).
\newblock
\showISSN{0360-0300}


\bibitem[Wang et~al\mbox{.}(2025)]%
        {NullSwap2025Wang[14]}
\bibfield{author}{\bibinfo{person}{Tianyi Wang}, \bibinfo{person}{Shuaicheng Niu}, \bibinfo{person}{Harry Cheng}, \bibinfo{person}{Xiao Zhang}, {and} \bibinfo{person}{Yinglong Wang}.} \bibinfo{year}{2025}\natexlab{}.
\newblock \showarticletitle{NullSwap: Proactive Identity Cloaking Against Deepfake Face Swapping}. In \bibinfo{booktitle}{\emph{Proceedings of the IEEE/CVF International Conference on Computer Vision}}.
\newblock


\bibitem[Woop et~al\mbox{.}(2011)]%
        {PeanoTexture2011[42]}
\bibfield{author}{\bibinfo{person}{Sven Woop}, \bibinfo{person}{Carsten Benthin}, {and} \bibinfo{person}{Ingo Wald}.} \bibinfo{year}{2011}\natexlab{}.
\newblock \showarticletitle{Efficient ray tracing of procedural textures}. In \bibinfo{booktitle}{\emph{Proceedings of the ACM SIGGRAPH Symposium on High Performance Graphics}}. \bibinfo{pages}{109--117}.
\newblock
\href{https://doi.org/10.1145/2018323.2018340}{doi:\nolinkurl{10.1145/2018323.2018340}}


\bibitem[Wu et~al\mbox{.}(2023)]%
        {SepMark[2]}
\bibfield{author}{\bibinfo{person}{Xiaoshuai Wu}, \bibinfo{person}{Xin Liao}, {and} \bibinfo{person}{Bo Ou}.} \bibinfo{year}{2023}\natexlab{}.
\newblock \showarticletitle{SepMark: Deep Separable Watermarking for Unified Source Tracing and Deepfake Detection}. In \bibinfo{booktitle}{\emph{Proceedings of the 31st ACM International Conference on Multimedia}}.
\newblock


\bibitem[Xu et~al\mbox{.}(2022)]%
        {UniFace[6]}
\bibfield{author}{\bibinfo{person}{Chao Xu}, \bibinfo{person}{Jiangning Zhang}, \bibinfo{person}{Yue Han}, \bibinfo{person}{Guanzhong Tian}, \bibinfo{person}{Xianfang Zeng}, \bibinfo{person}{Ying Tai}, \bibinfo{person}{Yabiao Wang}, \bibinfo{person}{Chengjie Wang}, {and} \bibinfo{person}{Yong Liu}.} \bibinfo{year}{2022}\natexlab{}.
\newblock \showarticletitle{Designing One Unified Framework for High-Fidelity Face Reenactment and Swapping}. In \bibinfo{booktitle}{\emph{European Conference on Computer Vision}}. \bibinfo{pages}{54--71}.
\newblock
\showISBNx{978-3-031-19784-0}


\bibitem[Yang et~al\mbox{.}(2021)]%
        {FaceGuard[23]}
\bibfield{author}{\bibinfo{person}{Yuankun Yang}, \bibinfo{person}{Chenyue Liang}, \bibinfo{person}{Hongyu He}, \bibinfo{person}{Xiaoyu Cao}, {and} \bibinfo{person}{Neil~Zhenqiang Gong}.} \bibinfo{year}{2021}\natexlab{}.
\newblock \bibinfo{title}{FaceGuard: Proactive Deepfake Detection}.
\newblock


\bibitem[Zhang et~al\mbox{.}(2024)]%
        {EditGuard[4]}
\bibfield{author}{\bibinfo{person}{Xuanyu Zhang}, \bibinfo{person}{Runyi Li}, \bibinfo{person}{Jiwen Yu}, \bibinfo{person}{Youmin Xu}, \bibinfo{person}{Weiqi Li}, {and} \bibinfo{person}{Jian Zhang}.} \bibinfo{year}{2024}\natexlab{}.
\newblock \showarticletitle{Editguard: Versatile image watermarking for tamper localization and copyright protection}. In \bibinfo{booktitle}{\emph{Proceedings of the IEEE/CVF Conference on Computer Vision and Pattern Recognition}}. \bibinfo{pages}{11964--11974}.
\newblock


\bibitem[Zhang et~al\mbox{.}(2025)]%
        {OmniGuard[37]}
\bibfield{author}{\bibinfo{person}{Xuanyu Zhang}, \bibinfo{person}{Zecheng Tang}, \bibinfo{person}{Zhipei Xu}, \bibinfo{person}{Runyi Li}, \bibinfo{person}{Youmin Xu}, \bibinfo{person}{Bin Chen}, \bibinfo{person}{Feng Gao}, {and} \bibinfo{person}{Jian Zhang}.} \bibinfo{year}{2025}\natexlab{}.
\newblock \showarticletitle{OmniGuard: Hybrid Manipulation Localization via Augmented Versatile Deep Image Watermarking}. In \bibinfo{booktitle}{\emph{Proceedings of the IEEE/CVF Conference on Computer Vision and Pattern Recognition (CVPR)}}. \bibinfo{pages}{3008--3018}.
\newblock


\bibitem[Zhao et~al\mbox{.}(2023b)]%
        {DiffSwap[8]}
\bibfield{author}{\bibinfo{person}{Wenliang Zhao}, \bibinfo{person}{Yongming Rao}, \bibinfo{person}{Weikang Shi}, \bibinfo{person}{Zuyan Liu}, \bibinfo{person}{Jie Zhou}, {and} \bibinfo{person}{Jiwen Lu}.} \bibinfo{year}{2023}\natexlab{b}.
\newblock \showarticletitle{DiffSwap: High-Fidelity and Controllable Face Swapping via 3D-Aware Masked Diffusion}. In \bibinfo{booktitle}{\emph{IEEE Conference on Computer Vision and Pattern Recognition}}. \bibinfo{pages}{8568--8577}.
\newblock


\bibitem[Zhao et~al\mbox{.}(2023a)]%
        {Zhao2023WACV[36]}
\bibfield{author}{\bibinfo{person}{Yuan Zhao}, \bibinfo{person}{Bo Liu}, \bibinfo{person}{Ming Ding}, \bibinfo{person}{Baoping Liu}, \bibinfo{person}{Tianqing Zhu}, {and} \bibinfo{person}{Xin Yu}.} \bibinfo{year}{2023}\natexlab{a}.
\newblock \showarticletitle{Proactive Deepfake Defence via Identity Watermarking}. In \bibinfo{booktitle}{\emph{2023 IEEE/CVF Winter Conference on Applications of Computer Vision}}. \bibinfo{pages}{4591--4600}.
\newblock
\href{https://doi.org/10.1109/WACV56688.2023.00458}{doi:\nolinkurl{10.1109/WACV56688.2023.00458}}


\bibitem[Zhu et~al\mbox{.}(2018)]%
        {HiDDeN[33]}
\bibfield{author}{\bibinfo{person}{Jiren Zhu}, \bibinfo{person}{Russell Kaplan}, \bibinfo{person}{Justin Johnson}, {and} \bibinfo{person}{Li Fei-Fei}.} \bibinfo{year}{2018}\natexlab{}.
\newblock \showarticletitle{HiDDeN: Hiding Data With Deep Networks}. In \bibinfo{booktitle}{\emph{European Conference on Computer Vision}}. \bibinfo{pages}{682--697}.
\newblock
\showISBNx{978-3-030-01267-0}


\end{thebibliography}

\clearpage

\begin{figure*}
  \includegraphics[width=\textwidth]{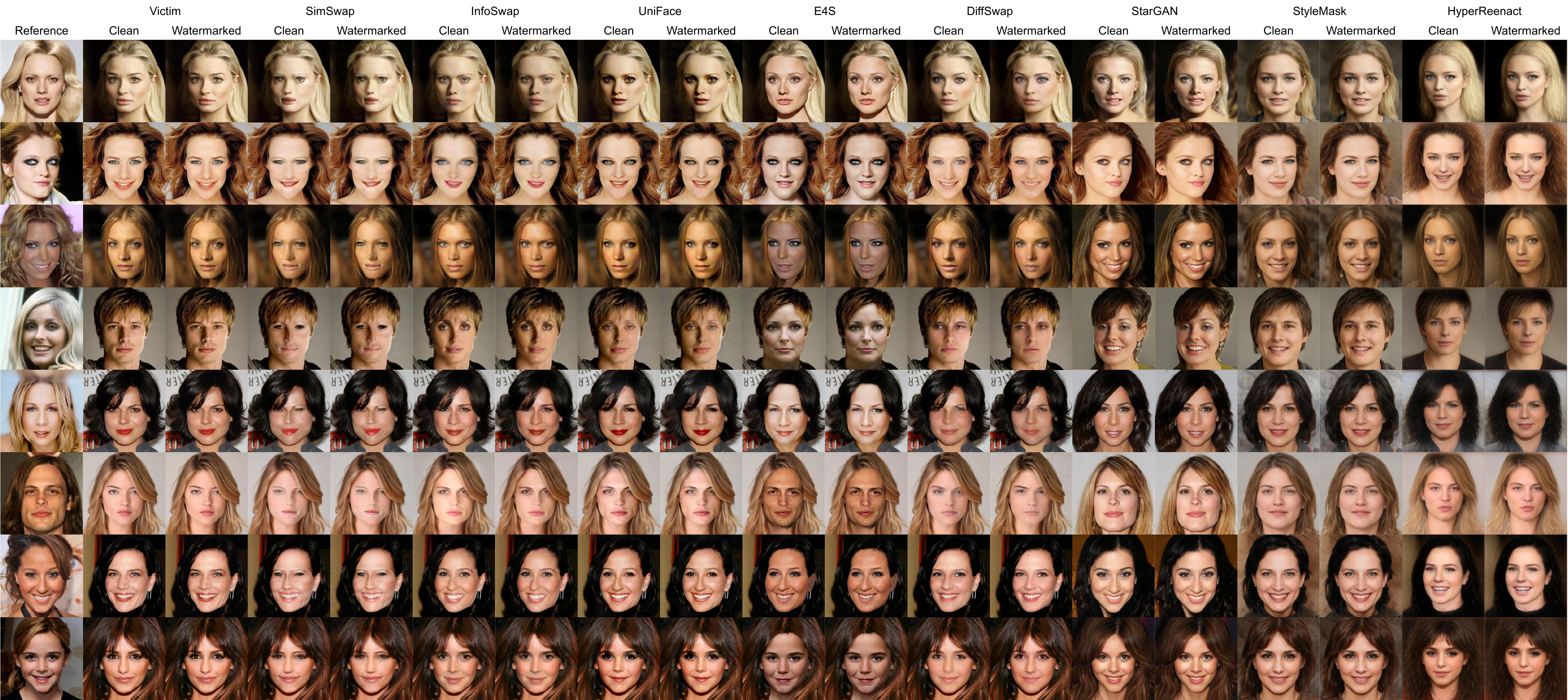}
  \caption{Deepfake results on raw and watermarked images. The watermark generally does not affect the synthetic results, which benefits the potential benign utilization of Deepfake manipulations. }
  \label{fig:teaser_all}
\end{figure*}

\appendix

\section{Extra Experimental Results}

In this section, we report extra experimental results in addition to those discussed in the main content.

\subsection{Visual Quality.}

In Figure~\ref{fig:teaser_all}, with regard to the reference faces, we applied each Deepfake manipulation to the clean and watermarked victim images in every two columns. In other words, we evaluated whether the embedded watermarks heavily affect the Deepfake manipulation results. It can be observed that, although trivial differences appear\footnote{DiffSwap~\cite{DiffSwap[8]} contains randomness upon every run, leading to differences even without watermarks. }, the Deepfake manipulations are generally unaffected, allowing potential benign utilization. 

\begin{figure}[t!]
  \includegraphics[width=\columnwidth]{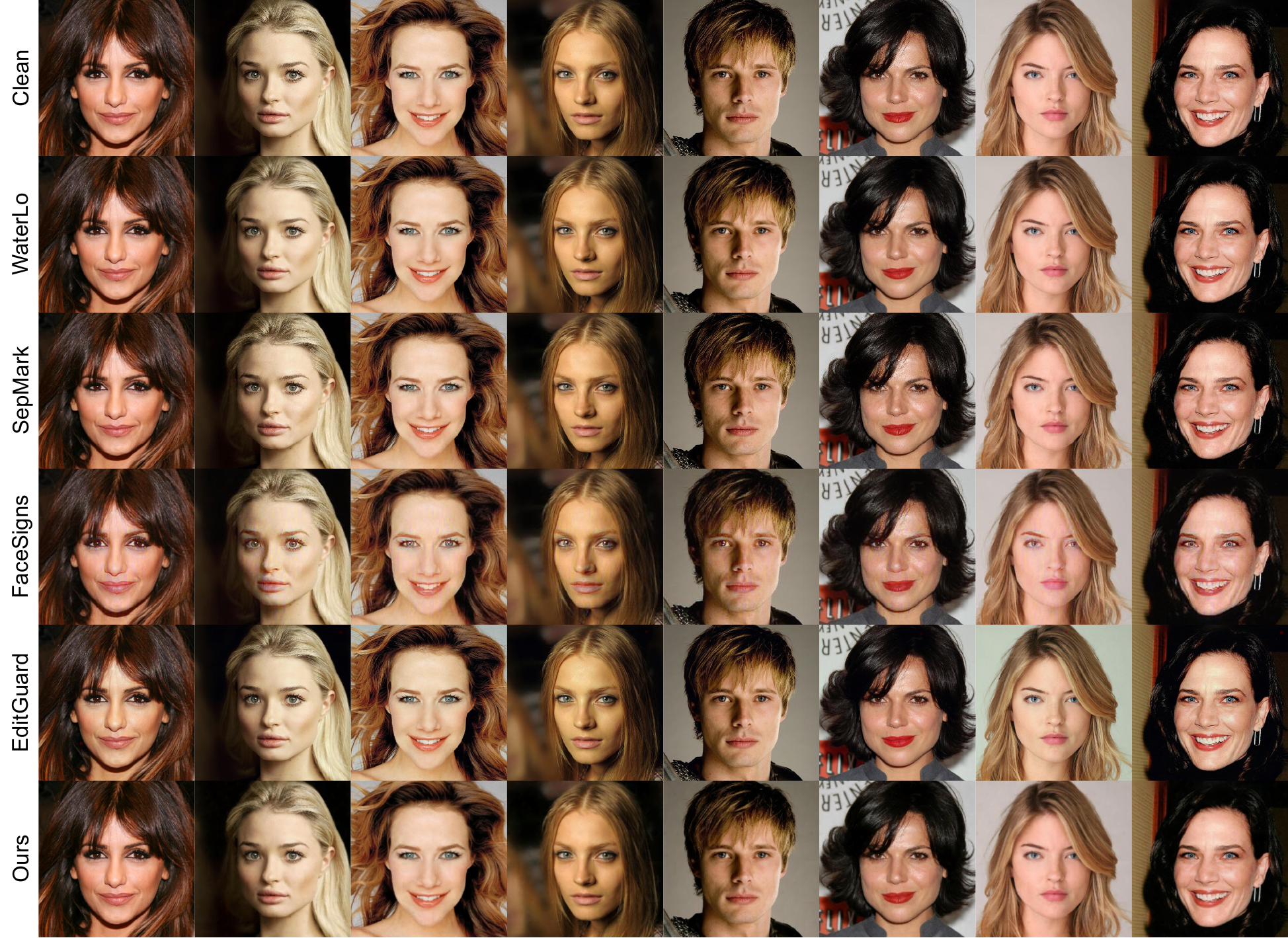}
  \caption{Watermarked images produced by each semi-fragile watermarking framework. }
  \label{fig:sup_watermarked_imgs}
\end{figure}

In Figure~\ref{fig:sup_watermarked_imgs}, we visualized some watermarked images produced by each semi-fragile watermarking framework. It can be observed that the results are generally consistent with the quantitative visual quality evaluation in the main content. In particular, WaterLo~\cite{WaterLo[1]} and EditGuard~\cite{EditGuard[4]} demonstrate outstanding PSNR values, with noise artifacts barely appearing in the images. On the other hand, obvious artifacts and distortions caused by the watermarks are displayed in rows 3 and 4. Additionally, mismatches in color and tone can be found in the row of EditGuard (column 7 being the most obvious), which matches the low SSIM values as listed in the main content.

\subsection{Cross-Dataset Evaluation on LFW}

\begin{table*}[t!]
\centering
\caption{Extra cross-dataset evaluation of watermarked recovery rate against common image operations on LFW. }
\begin{tabular}{lccccccc|c}
\toprule
Model & Resolution & Identity & Jpeg & Gaussian Noise & Gaussian Blur & Median Blur & Resize & Average$\uparrow$ \\
\midrule
SepMark~\cite{SepMark[2]} & 128 & 99.34\% & 96.69\% & 86.71\% & 99.61\% & 99.91\% & 99.74\% & 97.00\% \\
FractalForensics & 512 & 99.99\% & 99.73\% & 98.40\% & 99.97\% & 99.99\% & 99.78\% & 99.64\% \\
\bottomrule
\end{tabular}
\label{tab:sup_lfw_common}
\end{table*}

\begin{table*}[t!]
\centering
\caption{Extra cross-dataset evaluation of watermarked recovery rate against Deepfake manipulations on LFW. }
\resizebox{\textwidth}{!}{
\begin{tabular}{lccccccccc|c}
\toprule
Model & Resolution & SimSwap~\cite{SimSwap[15]} & InfoSwap~\cite{InfoSwap[5]} & UniFace~\cite{UniFace[6]} & E4S~\cite{E4S[7]} & DiffSwap~\cite{DiffSwap[8]} & StarGAN~\cite{StarGAN-V2[9]} & StyleMask~\cite{StyleMask[10]} & HyperReenact~\cite{HyperReenact[11]} & Average$\downarrow$ \\
\midrule
SepMark~\cite{SepMark[2]} & 128 & 97.72\% & 99.10\% & 53.29\% & 99.71\% & 50.19\% & 49.90\% & 49.68\% & 50.04\% & 68.69\% \\
FractalForensics & 512 & 61.32\% & 63.96\% & 27.62\% & 66.83\% & 89.97\% & 9.98\% & 10.00\% & 9.99\% & 42.46\% \\
\bottomrule
\end{tabular}
}
\label{tab:sup_lfw_deepfake}
\end{table*}

In the main content, only the 256 resolution setting is adopted for SepMark~\cite{SepMark[2]} and FractalForensics for simplicity. In this section, we supplement the 128 resolution for SepMark and 512 resolution for FractalForensics. As listed in Table~\ref{tab:sup_lfw_common} and Table~\ref{tab:sup_lfw_deepfake}, our proposed FractalForensics consistently performs with expected robustness against common image processing operations and fragility against Deepfake manipulations. Meanwhile, SepMark~\cite{SepMark[2]} generally behaves well at the 128 resolution except for a watermark recovery rate below 90\% for Gaussian Noise. Moreover, as a clear gap can be found between the robustness and fragility of our approach at the 512 resolution, the performance of SepMark when facing SimSwap~\cite{SimSwap[15]}, InfoSwap~\cite{InfoSwap[5]}, and E4S~\cite{E4S[7]} is unsatisfactory with watermark recovery rates as high as above 97\%. It is also worth noting that, for our proposed FractalForensics, since DiffSwap~\cite{DiffSwap[8]} modifies relatively fewer patches compared to other synthetic models and facial areas in LFW~\cite{LFW[48]} occupy fewer patches than those of CelebA-HQ~\cite{CelebAHQ[47]}, greater watermark recovery rates can be observed when tested against DiffSwap. However, this is reasonable and does not affect the gap between robustness and fragility. 

\begin{figure}[t!]
  \includegraphics[width=\columnwidth]{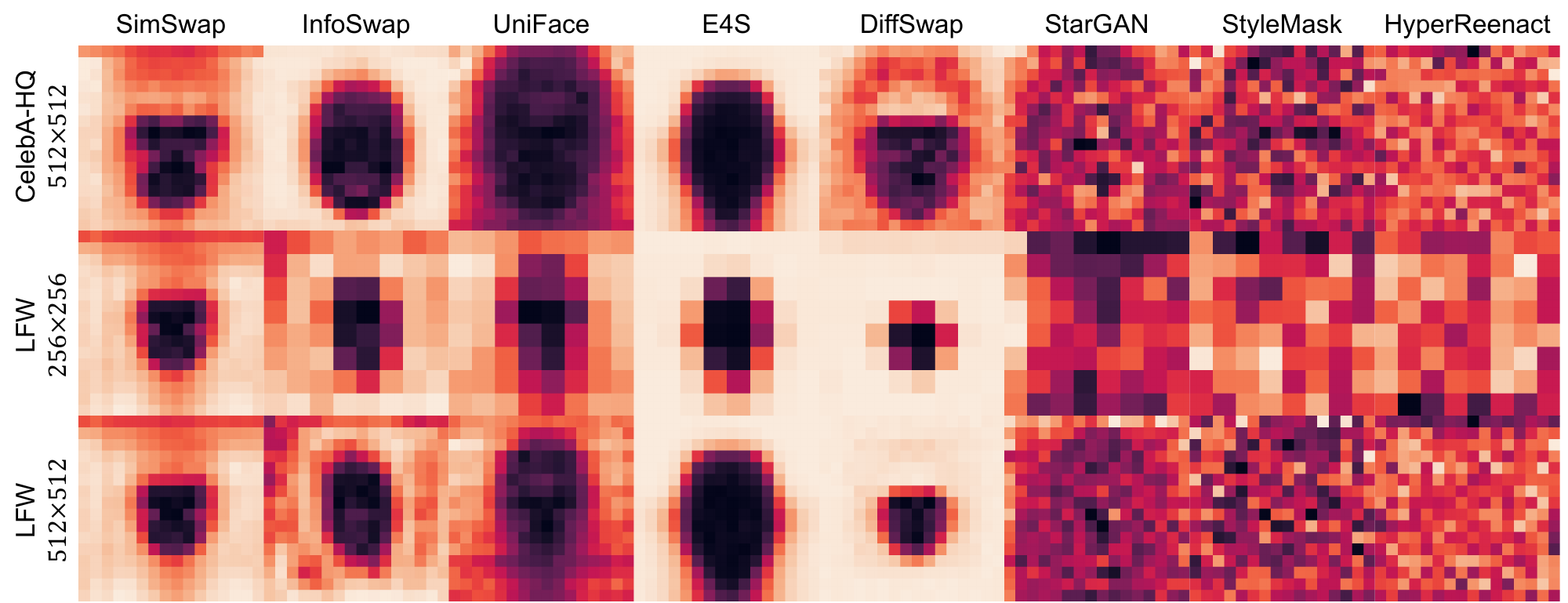}
  \caption{Cumulative localization visualizations of our FractalForensics against Deepfake manipulations on CelebA-HQ at 512 resolution and on LFW at 256 and 512 resolutions. }
  \label{fig:sup_localization_all}
\end{figure}

\begin{figure}
\centering
\includegraphics[width=\columnwidth]{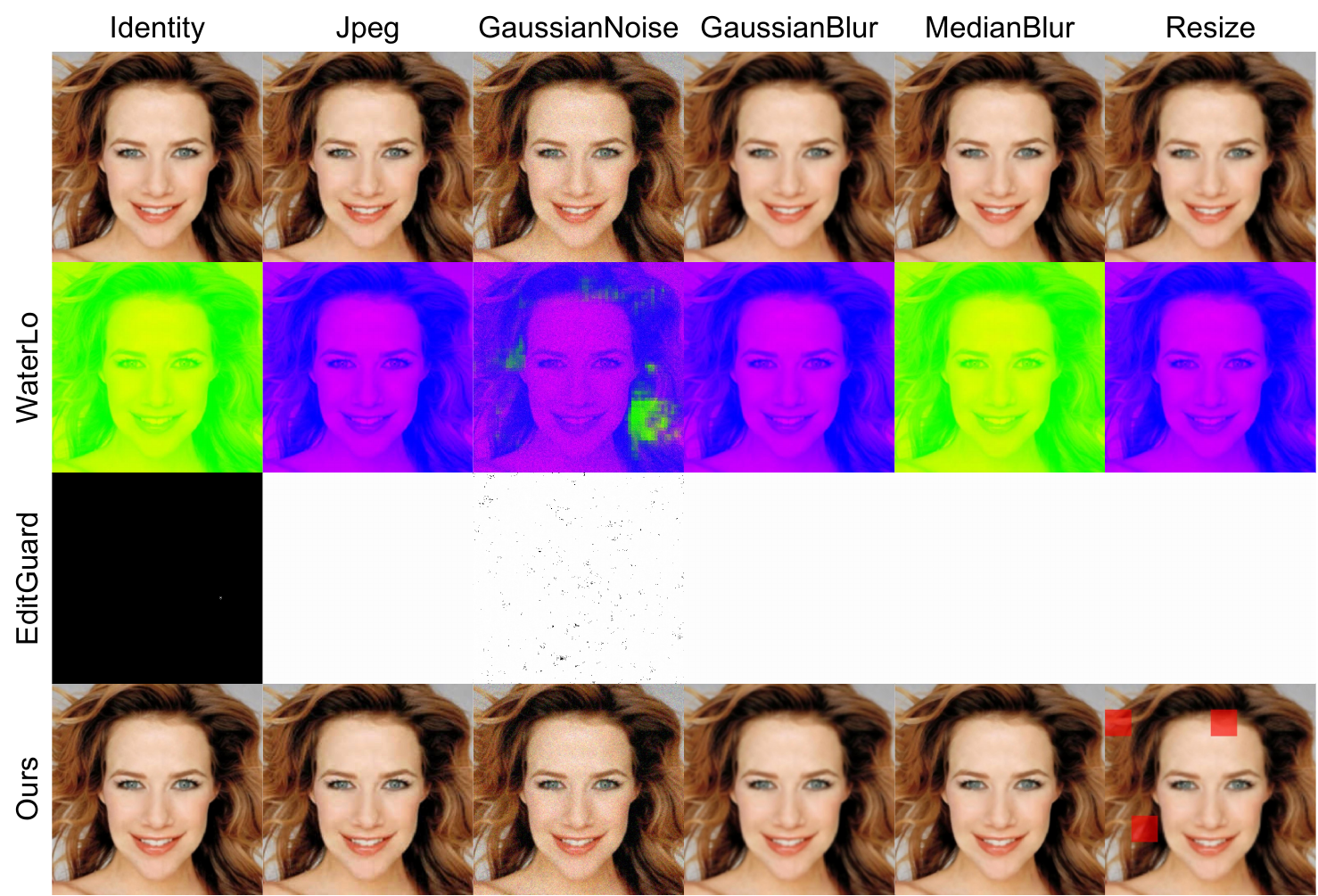}
\caption{Localization performance against common image operations. Fragility can be found for WaterLo~\cite{WaterLo[1]} and EditGuard~\cite{EditGuard[4]} for most operations. }
\label{fig:sup_localization_common}
\end{figure}

\subsection{Deepfake Localization}

In addition to the statistical watermark recovery rates for each experimental setting and the localization visualization for the 256 resolution on CelebA-HQ, in this section, we visualized the cumulative localization results of FractalForensics on CelebA-HQ~\cite{CelebAHQ[47]} at the 256 resolution and on LFW~\cite{LFW[48]} at 256 and 512 resolutions. In Figure~\ref{fig:sup_localization_all}, localized areas with highly mismatched watermarks are centered in the squares for face swapping models, which matches the functionality of face swapping algorithms. This makes sense since both CelebA-HQ and LFW contain facial images that have the faces aligned and mostly placed at the center while faces in LFW occupy smaller areas. Meanwhile, depending on the specific algorithm and model workflow, the amount of tampered patches varies for each swapping algorithm. As for the face reenactment models, since they edit head poses that affect the background area, most algorithms tend to smooth and blur the entire background area, leading to losing watermarks for those patches. In general, our proposed FractalForensics achieves promising and explainable localization for proactive Deepfake detection, regardless of datasets and image resolutions. 

Furthermore, we conducted detailed analyses regarding the localization results of WaterLo~\cite{WaterLo[1]} and EditGuard~\cite{EditGuard[4]} for comparisons with our FractalForensics. In Figure~\ref{fig:sup_localization_common}, WaterLo correctly identifies the image with no operation (Identity in column 1) as unmodified with a full overlay in green. However, it is highly affected by common operations and fails to maintain the desired watermark when facing Jpeg, Gaussian Noise, Gaussian Blur, and Resize. Similarly, EditGuard correctly localizes nothing in the first column, with a full dark localization result, but unexpectedly marks all common image operations as malicious edits with large white areas. 

\begin{figure}[t!]
\centering
\includegraphics[width=\columnwidth]{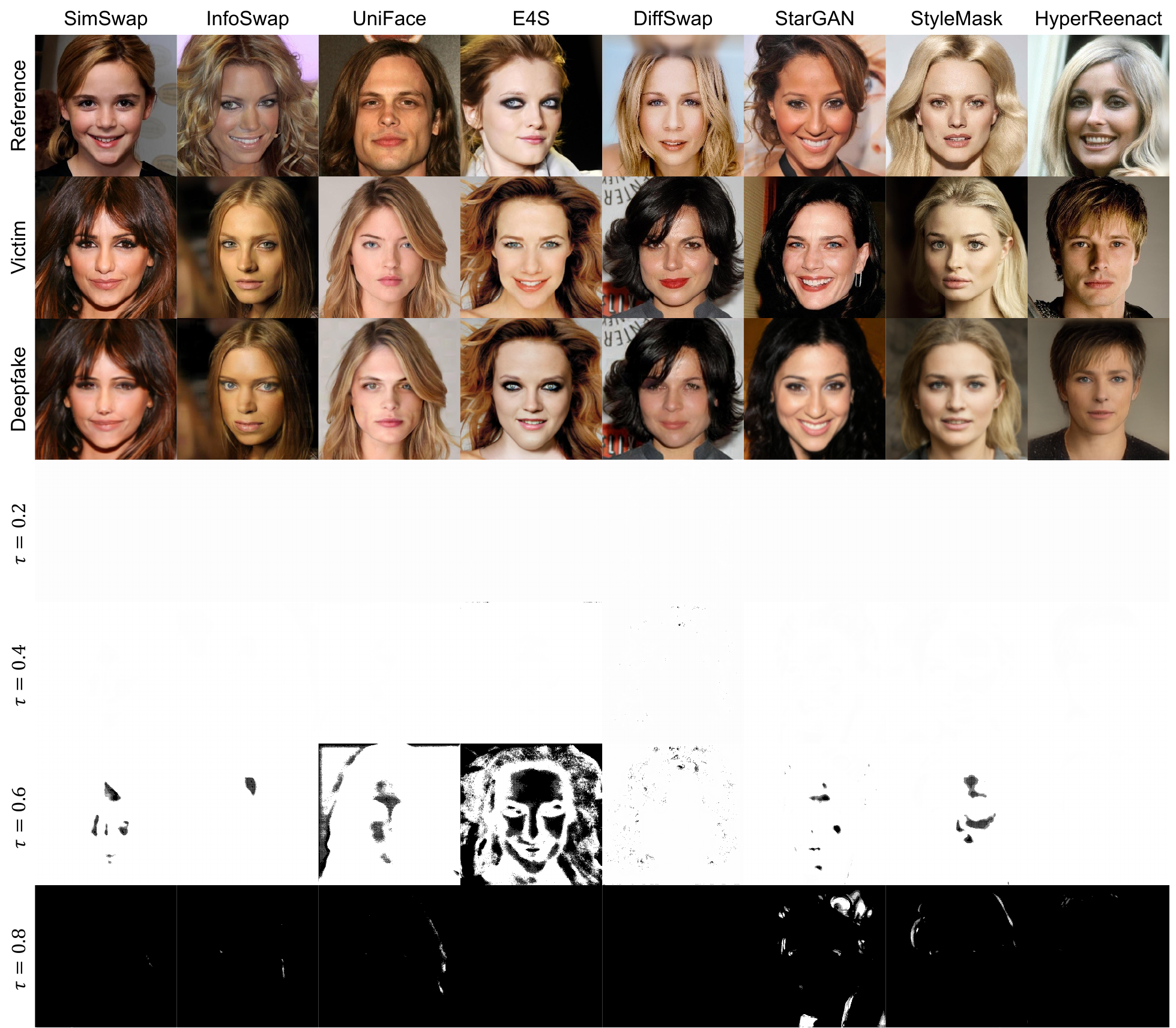}
\caption{Localization results of EditGuard as the threshold $\tau$ changes from 0.2 to 0.8. }
\label{fig:sup_editguard_localization}
\end{figure}

Additionally, the localization performance of EditGuard~\cite{EditGuard[4]} is controlled by a threshold $\tau$ that justifies the intensity of the watermark difference between recovered and ground-truth. In Figure~\ref{fig:sup_editguard_localization}, we visualized the localization results of EditGuard on Deepfake-manipulated images by gradually increasing the threshold $\tau$ from 0.2 up to 0.8. Note that white and black represent edited and untampered regions, respectively. It can be discovered that the white areas suddenly turn into black from $\tau=0.6$ to $\tau=0.8$. As a result, although the localization result varies against different Deepfake manipulations, for EditGuard, there lacks a reasonable localization result that clearly highlights the Deepfake-manipulated region rather than annotating the entire image as fake. 

\begin{table}[t!]
\centering
\caption{Watermark recovery performance against cropping areas of different sizes on CelebA-HQ. }
\resizebox{\columnwidth}{!}{
\begin{tabular}{lcccc}
\toprule
Num. Patches & Rec. Rate (bit) & Rec. Rate (patch) & Correctness & Rem. Portion \\
\midrule
$1 \times 1$ & 99.45\% & 98.79\% & 99.64\% & 98.44\% \\
$2 \times 2$ & 97.63\% & 94.79\% & 98.95\% & 93.75\% \\
$3 \times 3$ & 94.44\% & 87.89\% & 98.03\% & 85.94\% \\
$3 \times 4$ & 92.51\% & 83.74\% & 97.49\% & 81.25\% \\
$4 \times 4$ & 90.03\% & 78.32\% & 96.66\% & 75.00\% \\
$4 \times 5$ & 87.42\% & 72.59\% & 96.14\% & 68.75\% \\
$5 \times 5$ & 84.32\% & 65.76\% & 95.15\% & 60.94\% \\
Random & 95.09\% & 89.30\% & 98.19\% & -- \\
\bottomrule
\end{tabular}
}
\label{tab:sup_cropping_celebahq}
\end{table}

\subsection{Watermark Performance Against Cropping}

In the main content, it is proven that our proposed FractalForensics demonstrates side contributions with reasonable sensitivity regarding image cropping. In this section, in addition to the figure, in Table~\ref{tab:sup_cropping_celebahq} and Table~\ref{tab:sup_cropping_lfw}, we listed the detailed bit-wise watermark recovery rate, patch-wise watermark recovery rate, correctness, and remaining portion as the cropping size increases. In specific, the remaining portion (Rem. Portion) counts the percentage of non-cropped areas within the image, which provides a reference for evaluating the process that the patch-wise watermark recovery rate drops. In general, regardless of the in-dataset or cross-dataset settings, while consistently maintaining high correctness of watermark robustness and fragility, the patch-wise watermark recovery rate closely sticks with the remaining portion as the number of cropped patches increases. At the same time, the patch-wise watermark recovery rate is always slightly greater than the remaining portion, which demonstrates the trivial over-robustness of the watermarks of FractalForensics, possibly due to strong watermark correlations between locally adjacent patches.

\begin{table}[t!]
\centering
\caption{Watermark recovery performance against cropping areas of different sizes on LFW. }
\resizebox{\columnwidth}{!}{
\begin{tabular}{lcccc}
\toprule
Num. Patches & Rec. Rate (bit) & Rec. Rate (patch) & Correctness & Rem. Portion \\
\midrule
$1 \times 1$ & 99.42\% & 98.72\% & 99.59\% & 98.44\% \\
$2 \times 2$ & 97.56\% & 94.70\% & 98.93\% & 93.75\% \\
$3 \times 3$ & 94.40\% & 87.83\% & 97.99\% & 85.94\% \\
$3 \times 4$ & 92.45\% & 83.67\% & 97.47\% & 81.25\% \\
$4 \times 4$ & 89.89\% & 78.08\% & 96.77\% & 75.00\% \\
$4 \times 5$ & 87.36\% & 72.57\% & 96.05\% & 68.75\% \\
$5 \times 5$ & 84.15\% & 65.65\% & 95.17\% & 60.94\% \\
Random & 95.04\% & 89.26\% & 98.12\% & -- \\
\bottomrule
\end{tabular}
}
\label{tab:sup_cropping_lfw}
\end{table}

\begin{figure*}[t!]
    \centering
    \begin{subfigure}[t!]{0.32\textwidth}
        \includegraphics[width=\textwidth]{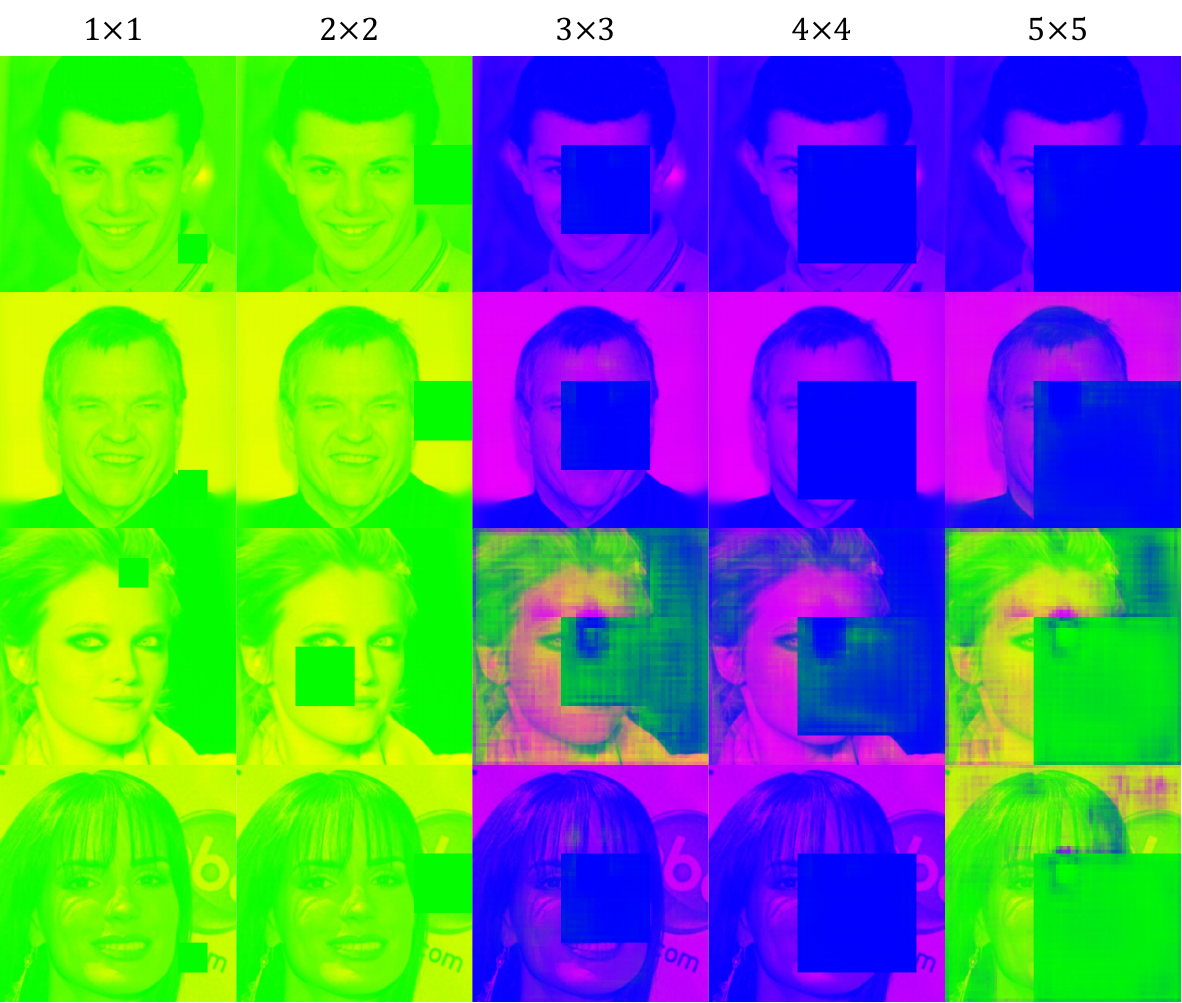}
        \caption{WaterLo~\cite{WaterLo[1]} against Cropping. }
    \end{subfigure}
    \hfill
    \begin{subfigure}[t!]{0.32\textwidth}
        \includegraphics[width=\textwidth]{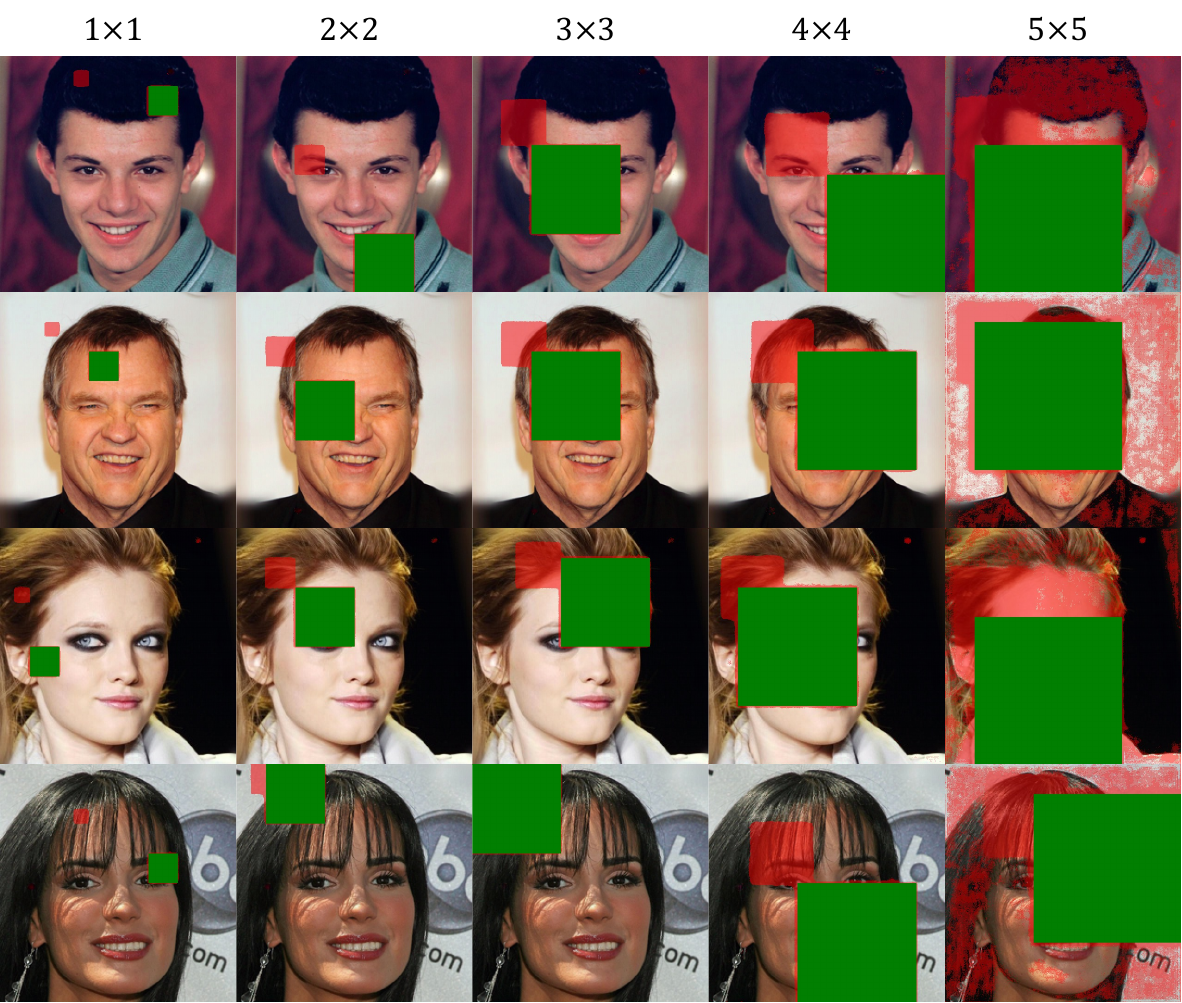}
        \caption{EditGuard~\cite{EditGuard[4]} against Cropping.}
    \end{subfigure}
    \hfill
    \begin{subfigure}[t!]{0.32\textwidth}
        \includegraphics[width=\textwidth]{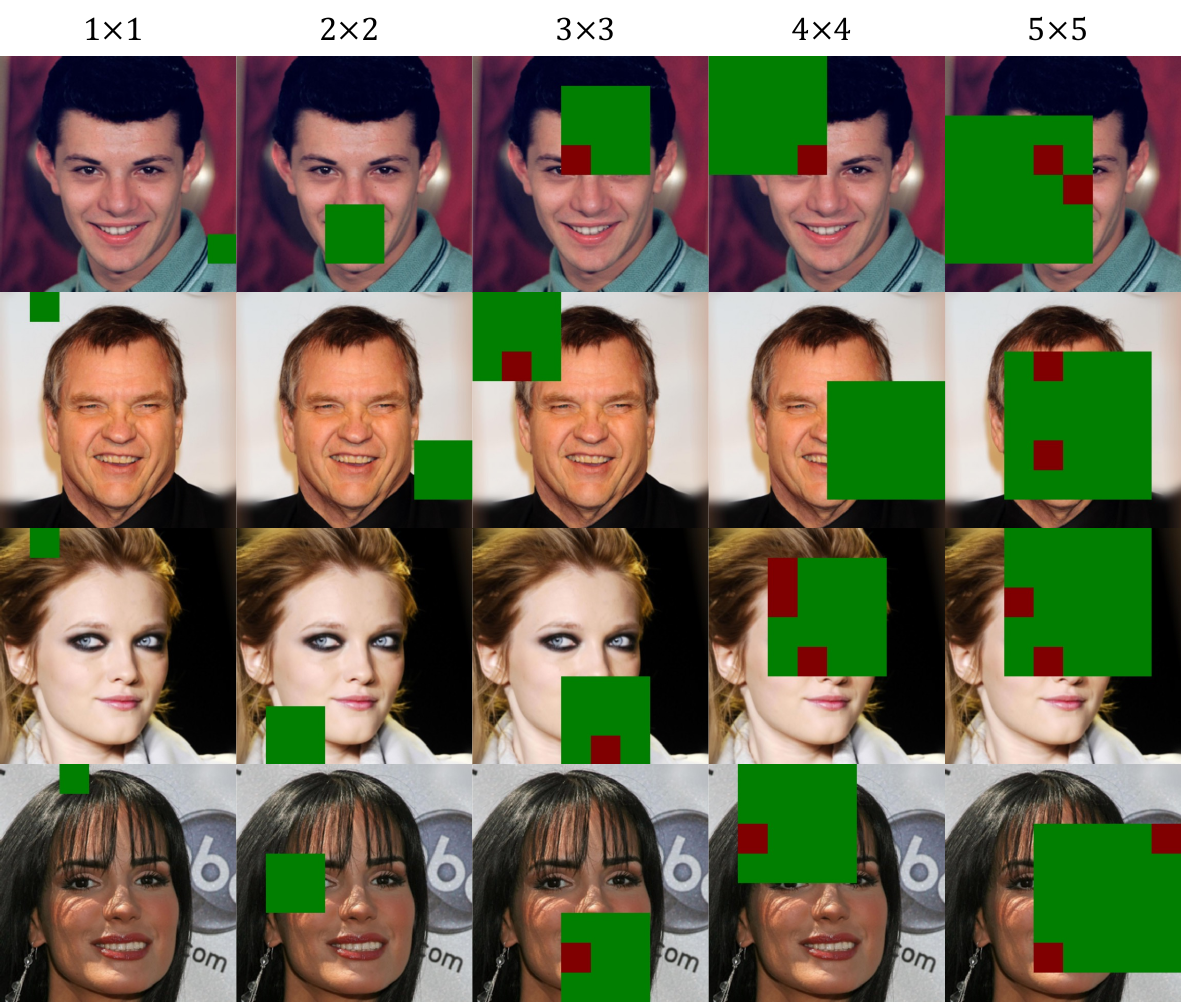}
        \caption{FractalForensics against Cropping.}
    \end{subfigure}
    \caption{Localization performance and correctness against Cropping. }
    \label{fig:sup_crop_local_compare}
\end{figure*}

In Figure~\ref{fig:sup_crop_local_compare}, we further visualized and compared the watermark localization performance when encountering different sizes of cropped patches. Since WaterLo~\cite{WaterLo[1]} provides end-to-end visualization of overlaying green and purple on the image to demonstrate real and fake areas, we mark the cropped patches as black to highlight the ground-truths. As for EditGuard~\cite{EditGuard[4]} and our proposed FractalForensics, besides highlighting the ground-truth cropped areas in black, we overlaid green and red for correct and incorrect localizations. In general, WaterLo is observed to have no ability to identify a small portion within the images but the entire images, with unrecognized cropped areas for the small number of cropped patches ($1 \times 1$ and $2 \times 2$) and recognized full images for larger numbers ($3 \times 3$, $4 \times 4$, and $5 \times 5$). On the other hand, EditGuard identifies extra areas in addition to the correctly localized patches, while ours sometimes appears to localize fewer cropped areas than the ground-truth. Additionally, the missed patches are mostly on the contours of the cropped areas, and this is because of the consecutive ConvBlocks and SEResBlocks that strengthen correlations between adjacent patches in the model architecture of FractalForensics. However, since moderate kernel sizes are applied, no correlation is found between patches that are far apart. 

\begin{table*}[t!]
\centering
\caption{Generalizability of Z-Order fractal watermarks against common image operations for watermark recovery rate. }
\resizebox{\textwidth}{!}{
\begin{tabular}{lccccccc|c|ccc}
\toprule
Dataset & Fractal & Identity & Jpeg & Gaussian Noise & Gaussian Blur & Median Blur & Resize & Average & PSNR & SSIM & LPIPS \\
\midrule
\multirow{2}{*}{CelebA-HQ} & Hilbert & 99.99\% & 99.97\% & 99.76\% & 99.57\% & 99.99\% & 98.03\% & 99.55\% & 35.925 & 0.974 & 0.013 \\
 & Z-Order & 99.99\% & 99.95\% & 99.76\% & 99.60\% & 99.87\% & 98.00\% & 99.53\% & 35.921 & 0.974 & 0.013 \\
 \midrule
\multirow{2}{*}{LFW} & Hilbert & 99.95\% & 99.53\% & 99.35\% & 98.97\% & 99.50\% & 97.16\% & 99.08\% & 35.7405 & 0.9695 & 0.0190 \\
 & Z-Order & 99.94\% & 99.55\% & 99.33\% & 98.95\% & 99.49\% & 97.16\% & 99.07\% & 35.7460 & 0.9695 & 0.0198 \\
\bottomrule
\end{tabular}
}
\label{tab:sup_z-order_common}
\end{table*}

\begin{table*}
\centering
\caption{Generalizability of Z-Order fractal watermarks against Deepfake manipulations for watermark recovery rate. }
\resizebox{\textwidth}{!}{
\begin{tabular}{lcccccccccc}
\toprule
Dataset & Fractals & SimSwap~\cite{SimSwap[15]} & InfoSwap~\cite{InfoSwap[5]} & UniFace~\cite{UniFace[6]} & E4S~\cite{E4S[7]} & DiffSwap~\cite{DiffSwap[8]} & StarGAN~\cite{StarGAN-V2[9]} & StyleMask~\cite{StyleMask[10]} & HyperReenact~\cite{HyperReenact[11]} & Average \\
\midrule
\multirow{2}{*}{CelebA-HQ} & Hilbert & 78.19\% & 75.12\% & 64.66\% & 69.19\% & 81.31\% & 9.92\% & 10.00\% & 9.98\% & 50.05\% \\
 & Z-Order & 78.25\% & 75.15\% & 47.64\% & 69.16\% & 81.80\% & 9.89\% & 10.09\% & 9.97\% & 47.74\% \\
\midrule
\multirow{2}{*}{LFW} & Hilbert & 74.87\% & 68.57\% & 51.49\% & 82.60\% & 91.48\% & 9.97\% & 10.00\% & 10.08\% & 49.88\% \\
& Z-Order & 74.95\% & 68.52\% & 51.35\% & 82.51\% & 92.41\% & 9.90\% & 9.97\% & 10.62\% & 49.95\% \\
\bottomrule
\end{tabular}
}
\label{tab:sup_z-order_fake}
\end{table*}

\section{Generalizability of the FractalForensics}

\subsection{Variations of Hilbert Curve}

\begin{figure}
    \centering
    \includegraphics[width=\columnwidth]{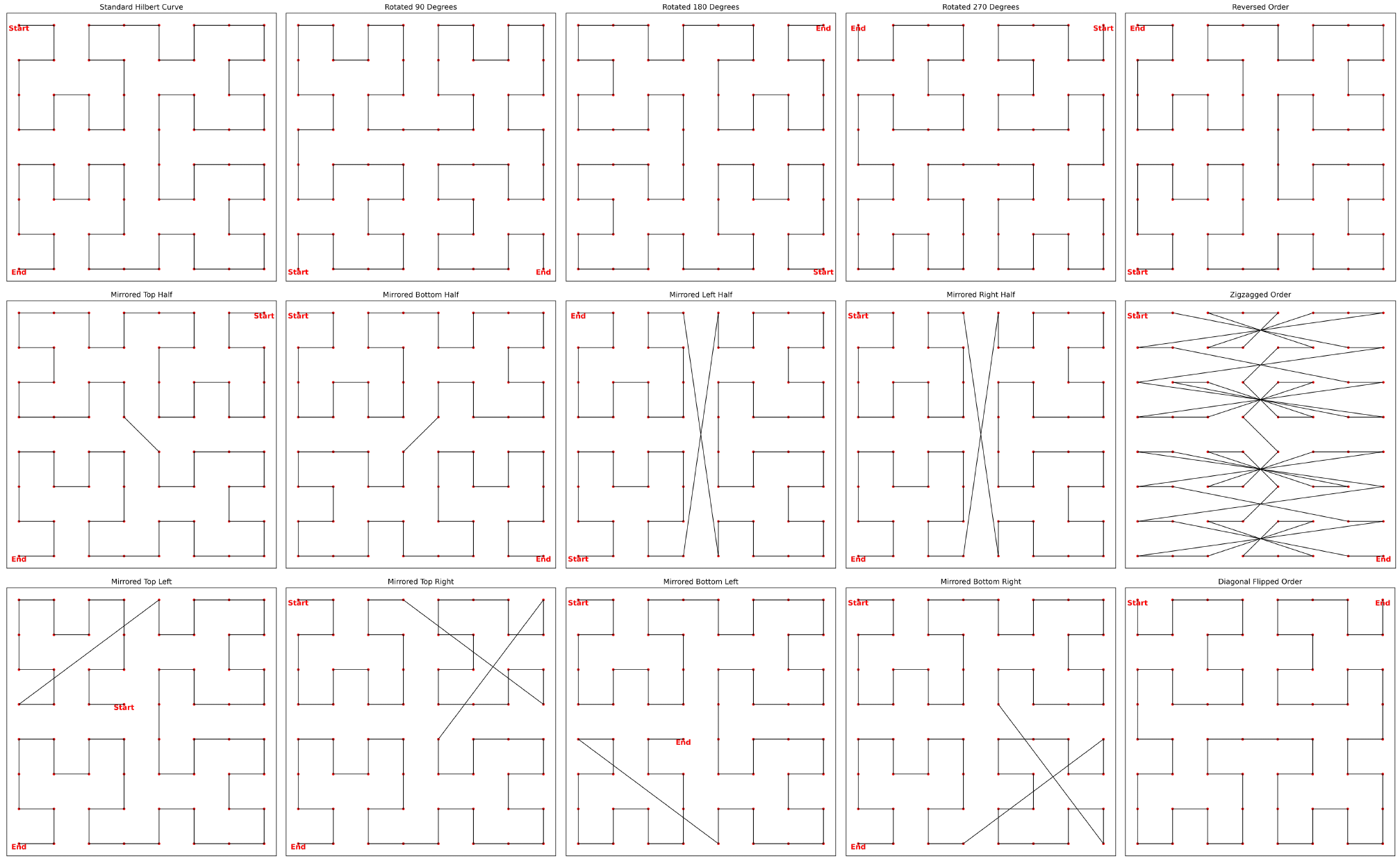}
    \caption{Individual variations of the standard Hilbert curve with respect to rotations, mirroring, and order modification. }
    \label{fig:sup_hilbert_curves}
\end{figure}

In the main paper content, the Hilbert curve is adopted as the standard base fractal shape for demonstrations and experiments. For safety and security purposes, we ensured diversity in the candidate watermark shapes by applying variations to the standard Hilbert curve, including 3 rotations, 8 mirroring, and 3 order modifications, leading to 144 unique combinations in total. To better explain why such variations help, in Figure~\ref{fig:sup_hilbert_curves}, we visualized the 14 individual variations in addition to the standard Hilbert curve. Note that although there is limited space such that we were not able to display all 144 unique fractal shapes, by looking at what is displayed, it can be easily implied that the combined variations are more complex, benefiting our goal of confidentially protecting the watermarks. 

\subsection{Generalization to Z-Order Curve}

\begin{figure}
    \centering
    \includegraphics[width=\columnwidth]{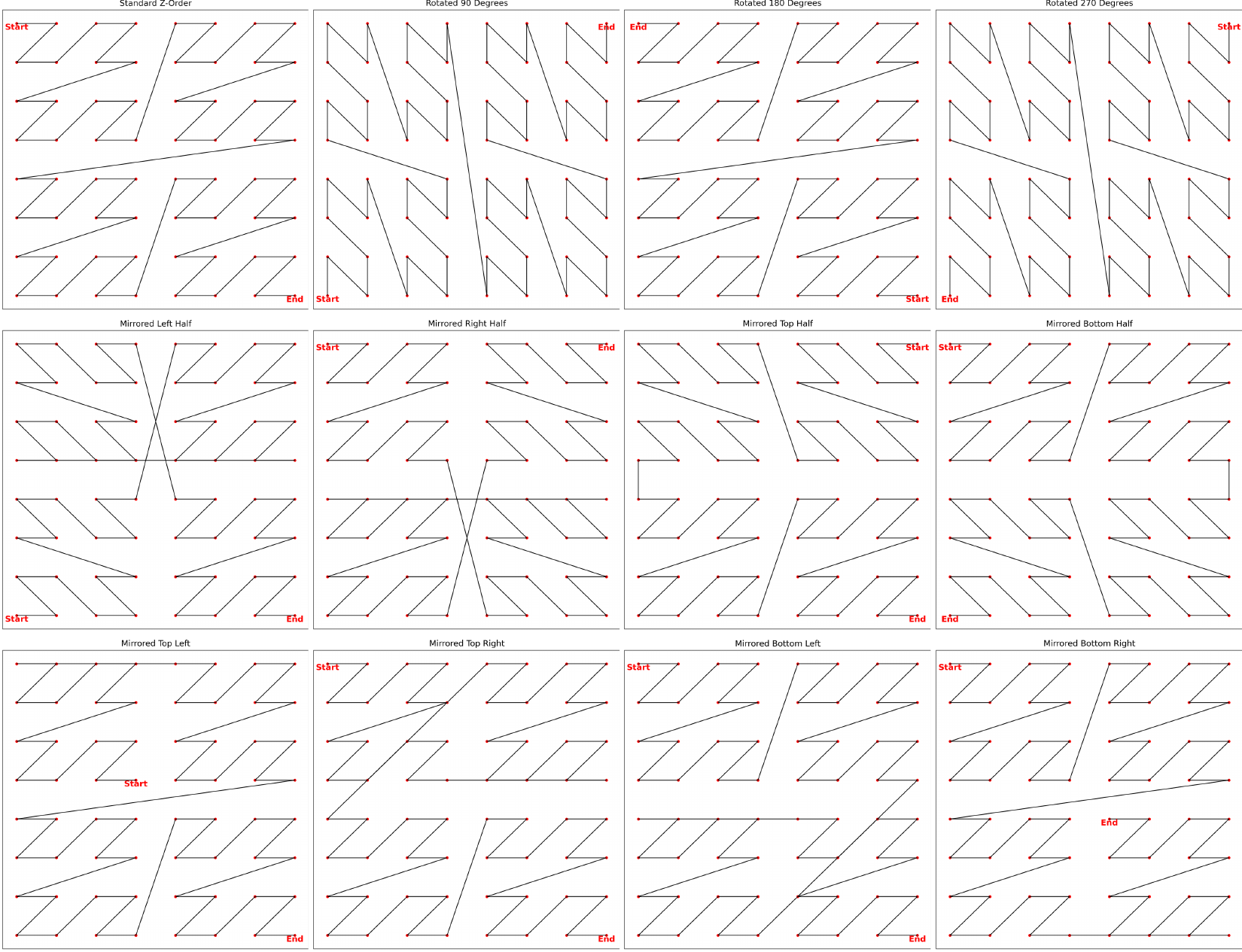}
    \caption{Individual variations of the standard Z-Order curve with respect to rotations, mirroring, and order modification. }
    \label{fig:sup_z-order_curves}
\end{figure}

In this study, the semi-fragile watermarking framework is trained based on watermarks that are variations of the standard Hilbert curve. However, in practice, a user may wish to select the type of fractal shapes. Therefore, in this section, we conducted watermark generation based on another fractal shape that properly fits our watermark sizes, namely the Z-Order curve~\cite{ZorderMemoryLayout2000[49]}. Due to the variation capacity, we applied the 3 rotations and 8 mirroring while omitting the order modifications, leading to 36 unique variation combinations. Again, due to limited space, we visualized the individual variations along with the standard Z-Order curve in Figure~\ref{fig:sup_z-order_curves} for better understanding. 

Upon preparing watermarks based on the Z-Order curves, we tested the proposed FractalForensics in terms of watermark recovery rate regarding robustness and fragility. The results are shown in Table~\ref{tab:sup_z-order_common} and Table~\ref{tab:sup_z-order_fake}. For the robustness against common image operations, it can be concluded that our FractalForensics is not affected by unseen fractal shapes although the average values for Z-Order curves, 99.53\% and 99.07\%, are slightly lower than those for Hilbert curves at 99.55\% and 99.08\%, respectively. On the other hand, the Z-Order-based watermarks perform regularly against Deepfake manipulations and there are still clear gaps between robustness and fragility. 

\section{Explanation on Watermark Confidentiality}

\begin{figure}
\centering
\includegraphics[width=\columnwidth]{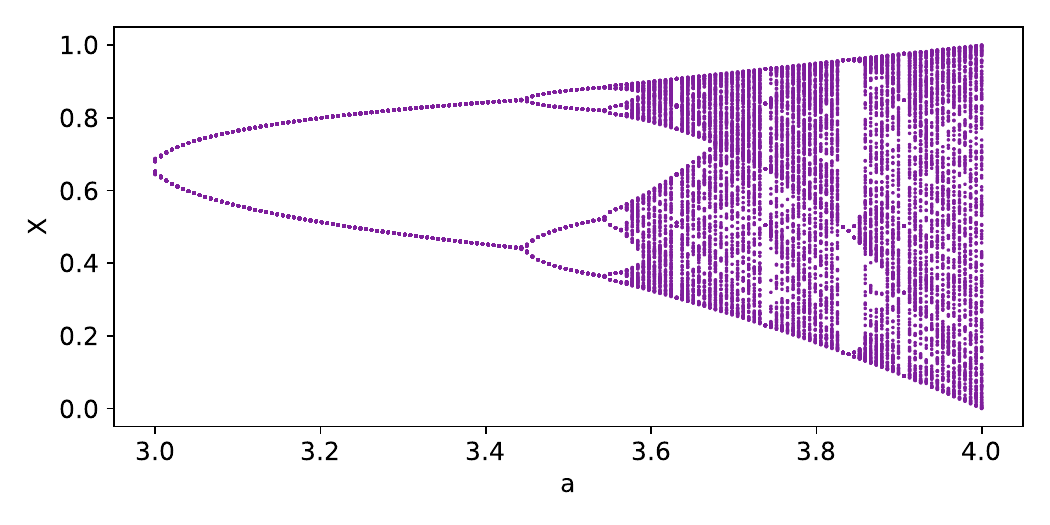}
\caption{Visualization of the bifurcation diagram for $a\in [3, 4]^\mathbb{R}$ with $x_0=0.1$. Dots with the same $a$ value display the values in the encrypted sequence. }
\label{fig:sup_chaotic_map}
\end{figure}

In this study, a chaotic encryption system is built to securely protect the watermarks against individuals who are not confidential. In this section, we demonstrated why the selected parameter ranges provide unpredictability and irreversibility of the one-way encryption. Regarding the logistic map 
\begin{equation}
x_i = ax_{i-1}(1-x_{i-1}),
\end{equation}
with user-selected $x_0 \in [0.1, 0.9]^\mathbb{R}$ and $a \in [3.7, 4.0)^\mathbb{R}$. The progression of the sequence of values is of no pattern, which can only be reproduced via the same set of parameters. In Figure~\ref{fig:sup_chaotic_map}, we plotted the bifurcation diagram of the logistic map with $x_0 = 0.1$ for the continuous range $a \in [3, 4]^\mathbb{R}$. Considering the restriction such that $k \in [100, 1000]^\mathbb{Z}$, we omitted $x_i$ for $i \in [0, 99]^\mathbb{Z}$, and plotted 200 consecutive $x_i$ values for each $a$ value by setting $k = 100$. It can be observed that the chaotic characteristic starts to occur after around $a = 3.6$, where $a \in [3.7, 4.0)^\mathbb{R}$ is a safe range for encryption. 

\clearpage

\end{document}